\documentclass[preprint,12pt]{elsarticle}

\usepackage{amssymb}
\usepackage{multirow}
\usepackage{makecell}
\usepackage{multicol}
\usepackage{textcomp}
\usepackage{threeparttable}
\usepackage{rotating}
\usepackage{xcolor}
\usepackage{soul}
\sethlcolor{yellow}
\usepackage{longtable}
\usepackage{makecell}
\usepackage{url}
\usepackage{adjustbox}

\usepackage{xcolor}
\usepackage{colortbl}
\definecolor{lightgray}{gray}{0.92}

\usepackage{caption}
\newcolumntype{Y}{>{\centering\arraybackslash}X}
\usepackage{algorithm,algcompatible,algpseudocode}
\usepackage{graphicx} 
\usepackage{array}
\newcolumntype{Y}{>{\raggedright\arraybackslash}X}
\usepackage{amsmath}
\usepackage{amsthm}
\usepackage{pifont}
\usepackage{booktabs}
\usepackage{tabularx}
\usepackage[most]{tcolorbox}
\usepackage[hidelinks]{hyperref}
\usepackage[pagewise]{lineno}

\begin{document}

\begin{frontmatter}

\title{Vision-Language Foundation Models for Comprehensive
Automated Pavement Condition Assessment}

\author[1]{Blessing Agyei Kyem}
\ead{blessing.agyeikyem@ndsu.edu}

\author[1]{Joshua Kofi Asamoah}
\ead{joshua.asamoah@ndsu.edu}

\author[2]{Anthony Dontoh}
\ead{adontoh@memphis.edu}

\author[1]{Armstrong Aboah\corref{cor1}}
\ead{armstrong.aboah@ndsu.edu}

\cortext[cor1]{Corresponding author}

\affiliation[1]{%
  organization={Department of Civil, Construction and Environmental Engineering, North Dakota State University},%
  city={Fargo},%
  postcode={58102},%
  state={ND},%
  country={USA}%
}

\affiliation[2]{%
  organization={Department of Civil, Construction and Environmental Engineering, University of  Memphis},%
  city={Memphis},%
  postcode={38152},%
  state={TN},%
  country={USA}%
}


\begin{abstract}
General-purpose vision-language models demonstrate strong performance in
everyday domains but struggle with specialized technical fields requiring precise terminology, structured reasoning, and adherence to engineering standards. This work addresses whether domain-specific instruction tuning can
enable comprehensive pavement condition assessment through vision-language
models. PaveInstruct, a dataset containing 278,889 image-instruction-response
pairs spanning 32 task types, was created by unifying annotations from nine
heterogeneous pavement datasets. PaveGPT, a pavement foundation model
trained on this dataset, was evaluated against state-of-the-art vision-language
models across perception, understanding, and reasoning tasks. Instruction
tuning transformed model capabilities, achieving improvements exceeding
20\% in spatial grounding, reasoning, and generation tasks while producing ASTM D6433-compliant outputs. These results enable transportation
agencies to deploy unified conversational assessment tools that replace multiple specialized systems, simplifying workflows and reducing technical expertise requirements. The approach establishes a pathway for developing
instruction-driven AI systems across infrastructure domains including bridge
inspection, railway maintenance, and building condition assessment.
\end{abstract}



\begin{keyword}
Vision-language models \sep Instruction tuning \sep Condition assessment \sep Infrastructure monitoring \sep Multimodal learning \sep Spatial grounding \sep Foundation models \sep ASTM D6433
\end{keyword}

\end{frontmatter}



\section{Introduction}
\label{intro}
Recent advances in large open-source vision-language models (VLMs), including QwenVL \cite{Qwen-VL,Qwen2-VL,Qwen2.5-VL}, LLaVA \cite{liu2023llava,liu2023improvedllava} have demonstrated strong capability in multimodal perception and language-guided reasoning tasks. While these models exhibit notable performance in open-domain environments, their direct application to specialized technical fields remains limited. Although proprietary models such as ChatGPT, Gemini, and Grok demonstrate strong multimodal capabilities in some technical fields, their limited transparency, inability to fine-tune, and data privacy concerns make them unsuitable for infrastructure assessment where regulatory compliance and reproducibility are essential. In high-precision domains such as medicine and autonomous driving, general-purpose VLMs have shown difficulty reasoning over domain-specific semantics, adhering to specialized terminology, and following structured expert protocols. These limitations have motivated the creation of domain-tailored instruction datasets such as Path-VQA \cite{he2020pathvqa} and VQA-RAD \cite{lau2018vqarad} for clinical imaging, and nuScenes-QA \cite{qian2023nuscenesqa}, DriveLM \cite{sima2024drivelm}, and BDD-X \cite{kim2018textual} for autonomous driving, leading to dedicated VLMs such as LLaVA-Med \cite{LLaVA_Med} and Med-PaLM \cite{singhal2022med_palm}. These developments underscore a broader conclusion: \textit{general-purpose VLMs are insufficient for expert-level reasoning in domain-critical applications, and performance gains require domain-aligned instruction datasets and supervision strategies.}

Pavement condition assessment presents similar challenges because it involves fine-grained distress identification, severity quantification aligned with engineering standards, precise spatial localization, and structured reporting to support maintenance decision-making. Existing instruction-tuning datasets such as LAION-5B \cite{schuhmann2022laion5b}, Conceptual Captions \cite{sharma2018conceptualcaptions}, COCO Captions \cite{chen2015mscococaptions}, Visual Genome \cite{krishna2017visualgenome}, and LLaVA-Instruct-150K \cite{liu2023llavainstruct150k} which are used to train these general-purpose VLMs contain minimal infrastructure content and do not encode pavement engineering terminology or reasoning processes. Consequently, current VLMs often misinterpret pavement distresses, provide nonspecific responses, and fail to follow standardized pavement evaluation procedures especially when tested in zeroshot scenarios. Although existing pavement datasets such as CrackSeg9k \cite{Kulkarni2022CrackSeg9k}, DeepCrack \cite{liu2019deepcrack}, Crack500 \cite{CRACK500}, Pavementscapes \cite{tong2022pavementscapes}, SVRDD \cite{SVRDD}, PaveDistress \cite{PaveDistress}, and recent captioning datasets \cite{Majidifard,Kyem2024PaveCapTF} have advanced detection, segmentation, and classification, these datasets are useful for unimodal vision-only tasks and do not provide text or instruction-response supervision required for multimodal technical reasoning, step-wise PCI estimation, or ASTM standards-compliant distress communication. Thus, while some recent works have begun experimenting with VLMs for pavement analysis, they remain constrained by the lack of instruction-grounded, standards-aligned datasets.

Recent work has applied VLMs to pavement tasks like zero-shot crack detection and few-shot damage assessment, revealing their potential \cite{ZHANG2025106389}. RoadBench \cite{roadbench}, for instance, offers a benchmark of synthetically captioned road images and introduces RoadCLIP, a non-generative, dual-encoder CLIP model for zero-shot classification and retrieval. However, these efforts face key limitations. First, they rely on prompting general-purpose models rather than fine-tuning domain-specific ones. Second, the captions in the dataset lack instruction-following, conversational structure and are not aligned with ASTM D6433 standards and guidelines. In addition, the models focus on narrow tasks such as classification or semantic localization. As a result, they fall short of the structured reasoning and instruction-following required for comprehensive pavement assessment, including distress detection, severity rating, PCI estimation, and maintenance suggestions. Bridging this gap requires instruction datasets grounded in domain standards and specialized multimodal models capable of end-to-end reasoning across all pavement tasks.

To address this gap, this study introduces \textbf{PaveInstruct}, a unique multimodal instruction-following dataset for pavement condition assessment, and \textbf{PaveGPT}, a domain-specialized foundation model trained on this dataset. PaveInstruct integrates pavement imagery with engineering-aligned prompts and structured responses covering distress identification and localization, ASTM-based severity assessment, chain-of-thought PCI estimation, formatted condition reporting, and maintenance recommendations. By explicitly aligning model supervision with pavement engineering workflows, PaveGPT enables technical dialogue, evidence-grounded reasoning, and standards-compliant interpretation of pavement conditions. This work establishes the foundation for instruction-driven pavement intelligence and introduces a unique language-native model designed for automated pavement evaluation and decision support.

The main contributions of the proposed approach are summarized below:
\begin{itemize}
\item PaveInstruct, a comprehensive instruction-following dataset containing 278,889 image-instruction response pairs spanning 32 task types across five major categories, is introduced and will be made publicly available for research purposes to advance vision-language models in infrastructure domains.

\item A systematic pipeline for integrating heterogeneous pavement datasets is developed, addressing annotation format unification, coordinate system harmonization, and task-specific instruction generation while preserving semantic richness and engineering validity across nine diverse data sources.

\item PaveGPT, a domain-specialized vision-language foundation model, is presented and demonstrates strong performance across perception, understanding, and reasoning tasks while maintaining computational efficiency suitable for practical deployment in pavement management systems.

\item Comprehensive empirical evidence is provided showing that domain-specific instruction tuning transforms general-purpose VLMs into capable pavement assessment tools, achieving improvements exceeding 20\% in spatial grounding, reasoning, and generation tasks, with consistent gains across different model architectures and sizes.
\end{itemize}



\section{Related Works}
\label{lit_review}
This section reviews existing pavement datasets and instruction-following multimodal datasets to establish the gap that PaveInstruct addresses. While pavement datasets have advanced distress detection through classification, detection, and segmentation tasks, they lack the natural language supervision required for training conversational assessment models, which instruction-following datasets have successfully enabled in domains such as medicine and autonomous driving.

\subsection{Existing Pavement datasets}
\label{existing_datasets_lit_review}
Numerous pavement datasets have been developed for distress detection, primarily targeting single computer vision tasks with fixed annotation formats. This section reviews classification, detection, and segmentation datasets, identifying their limitations for vision-language model training.

\textbf{Classification Datasets.} Early efforts produced image-level crack datasets to train classifiers distinguishing cracked vs. intact surfaces. For example, Özgenel et al. \cite{CCI4C} compiled the Concrete Crack Images for Classification dataset with 40,000 227×227 pixel images (20k with cracks, 20k without) derived from concrete surfaces on a university campus data. Similarly, Maguire et al. \cite{sdnet} released the SDNET2018 dataset containing 56,000 annotated images of concrete cracks and non-cracks on bridge decks, walls, and pavements. These large image-level datasets provided plentiful data for training deep classifiers, but they only offer coarse labels (crack present or not) without any spatial localization of distress within the image.

\noindent\textbf{Object Detection Datasets.} To enable spatial localization of pavement defects, several works have provided bounding-box annotations. Eisenbach et al. \cite{gaps_v1} introduced the German Asphalt Pavement Distress (GAPs) dataset, using vehicle-mounted cameras to collect road images labeled with boxes for six types of distresses. This was later extended by Stricker et al. \cite{gaps_v2} with an expanded GAPs dataset (GAPs v2) that increased the number of images and improved label quality. In Japan, Maeda et al. \cite{rdd2018} organized the Road Damage Dataset 2018 (RDD2018) with 9,053 roadway images and bounding-box annotations for common road damages. The RDD series has since grown: RDD2020 \cite{RDD2022}  included 26,336 images from Japan, India, and Chile, and the latest RDD2022 spans 47,420 images from six countries (adding the US, Norway, and China) labeled across four distress categories (longitudinal, transverse, alligator cracks, and potholes). More recently, Yang et al. \cite{PaveTrack} published PaveTrack, a large-scale two-part dataset: pavement images with multi-class distress bounding boxes for object detection and other set of images for tracking the temporal evolution of cracks and potholes. Likewise, Ren et al. \cite{SVRDD} developed the Street View Road Damage Dataset (SVRDD) using 8,000 panoramas from Baidu Street View, marking over 20,000 instances of pavement damage with bounding boxes. These detection-focused datasets substantially increased scale and diversity, covering multiple distress types and scenes. However, their annotations remain limited to predefined categories and do not capture fine-grained pixel details or any textual descriptions. 

\noindent\textbf{Segmentation Datasets}. For pixel-precise delineation of cracks, a variety of segmentation datasets have been developed. One early example is CrackTree260 by Zou et al. \cite{cracktree260}, which provided 260 images with cracks manually outlined. Another is the CrackForest dataset (CFD) introduced by Shi et al. \cite{crackforest}, comprising 118 road images with ground-truth binary masks for cracks. In the deep learning era, larger segmentation benchmarks emerged: Yang et al. \cite{CRACK500} compiled the Crack500 dataset with 500 pavement images, each expertly annotated at the pixel level. This dataset includes four common crack types (alligator, longitudinal, transverse, and block cracking) and presents realistic challenges like shadows and complex backgrounds. Similarly, Liu et al. \cite{zou2018deepcrack} released the DeepCrack dataset, which contains 537 high-resolution images of concrete and asphalt surfaces with finely labeled crack masks. To facilitate benchmarking, recent work has even combined multiple segmentation datasets. For instance, Kulkarni et al. \cite{crackseg9k} aggregated several sources into the CrackSeg9k collection (about 9,000 images) by unifying their annotations and addressing inconsistencies. Overall, segmentation datasets offer precise localization of distress, but each is typically focused on a narrow defect type (primarily cracks) and provides no semantic description beyond the mask itself. 

\noindent\textbf{PCI Datasets}. In contrast to segmentation datasets, fewer public datasets provide pavement images paired with PCI labels. One notable example is the PCIer dataset \cite{PCIer}, which contains pavement images categorized into color-coded condition ranges that correspond to PCI intervals, enabling learning-based condition assessment from visual inputs. More recently, the DSPS24 dataset \cite{DSPS} includes pavement surface images annotated with PCI scores and severity information, supporting supervised PCI estimation and condition classification. These datasets demonstrate that image-level PCI supervision is feasible and has begun to support learning-based condition prediction. However, their scale and diversity remain limited when compared to detection and segmentation benchmarks, and they typically provide coarse condition labels rather than fine-grained, distress-level reasoning. 

\subsection{Instruction-following multi-modal datasets}
Instruction-following datasets provide paired examples of multimodal inputs (e.g. images) with natural language instructions and responses. They enable VLMs to interpret questions, follow commands, and engage in interactive dialogue across tasks. These datasets have proven critical for training general-purpose multimodal assistants, as they improve zero-shot reasoning and align models with user intent. 

\noindent\textbf{General-Domain Instruction Datasets.} General purpose instruction datasets have been developed to help train multimodal LLMs to work across a variety of tasks. For instance, Liu et al. \cite{llava_instruct} introduced LLaVA-Instruct , using GPT-4 to generate 158K image-based instruction-response pairs for training VLMs. This approach produced one of the first multimodal instruction-following datasets and yielded the LLaVA assistant capable of open-ended image descriptions and question-answer pairs. Subsequent efforts scaled up both data and diversity. Chen et al. \cite{ShareGPT4V} presented ShareGPT4V, a resource of 1.2 million high-detail image captions created with GPT-4 Vision. The ShareGPT4V data covers broad visual concepts (objects, spatial relations, aesthetics) and significantly improved fine-tuning of VLMs on general benchmarks. Another notable work is M3IT by Li et al. \cite{li2023m3it}, a multilingual multimodal instruction corpus spanning 40 tasks with 2.4 million vision-text instances and queries translated into 80 languages. Together, these large-scale datasets enable general VLMs to follow open-ended instructions across everyday images. However, they remain focused on common domains and may lack specialized expertise such as in medicine and autonomous driving. 

\noindent\textbf{Domain-Specific Instruction Datasets.} Specialized domains have developed instruction-following datasets that inject expert knowledge into VLM training. In medical imaging, He et al. \cite{he2020pathvqa} created PathVQA with 4,998 pathology images and 32,799 clinical question-answer pairs, while Lau et al. \cite{lau2018vqarad} introduced VQA-RAD containing 315 radiology images with 3,515 clinician-written QA pairs \cite{lau2018vqarad}. These datasets enabled models such as LLaVA-Med \cite{LLaVA_Med} and Med-Flamingo \cite{Med_Flamingo}, which can reason through medical images using clinical terminology and generate diagnostic reasoning. In autonomous driving, Qian et al. \cite{qian2023nuscenesqa} developed NuScenes-QA with 34,000 street scenes and 460,000 question-answer pairs covering spatial reasoning and safety conditions. Kim et al. \cite{kim2018textual} also contributed BDD-X with 7,000 driving videos and human-written action explanations, while Deruyttere et al. \cite{talk2car} presented Talk2Car containing 12,000 natural language commands for vehicle control. These datasets enabled DriveVLM \cite{drivevlm}, a multimodal LLM which performs scene description and chain-of-thought navigation reasoning. Beyond healthcare and transportation, Lobry et al. \cite{RSVQA} introduced RSVQA for satellite image analysis using geographic metadata, while datasets such as ScienceQA \cite{ScienceQA} and AI2D \cite{AI2D} enable scientific diagram interpretation. Similarly, datasets such as DocVQA \cite{DocVQA} and InfographicVQA \cite{InfographicVQA} have enabled document understanding tasks. These domain-specific datasets share common characteristics: technical terminology, spatial reasoning requirements, and expert-level decision protocols. The consistent pattern across domains demonstrates that instruction-following datasets enable specialized VLMs with practical expert-assistant capabilities, yet civil infrastructure and pavement assessment remain conspicuously absent from this paradigm.

\section{Methodology}
This section describes the creation of PaveInstruct and the development of PaveGPT for instruction-driven pavement assessment. The source datasets and systematic pipeline for generating instruction-response pairs are presented first. The model architecture, training procedure, and comprehensive evaluation framework spanning perception, understanding, and reasoning tasks are then introduced.

\subsection{Source Datasets}
Table \ref{tab:raw_datasets} shows the datasets with raw annotations that served as our foundation to create the PaveInstruct dataset. The PaveInstruct dataset is built from several raw pavement datasets that contain original annotations in formats such as bounding boxes, segmentation masks, severity labels, and numeric PCI scores. These datasets were not created for instruction-following tasks, but they form the base from which instruction-response pairs were generated. Together, they cover a wide range of distress types, image perspectives, locations, weather scenarios, and annotation styles. 

Many of these datasets focus on spatial localization of pavement distresses. For example, the PID dataset includes bounding boxes for block cracks, transverse cracks, potholes, and even sealed variants such as sealed reflective and sealed transverse cracks. PaveTrack dataset adds patched potholes and clay-patched cracks, making it possible to distinguish between original distresses and repaired surfaces. These detailed labels create a strong foundation for generating instructions that reflect practical maintenance scenarios.

Other datasets contribute pixel-level segmentation masks and distress severity information. DSPS23 is a key example because it includes segmentation masks along with severity levels such as low, medium, and high. This supports instruction creation that aligns with standardized severity assessment. UAV-PDD2023, SVRDD, and UAPD add further diversity by providing annotated images from top-down UAV views and front-facing street-level perspectives, which broadens the range of visual conditions used to generate instructions.

Additional datasets supply pavement condition ratings that support PCI-focused instruction generation. DSPS24 provides raw numeric PCI scores from 0 to 100, which are suitable for general condition of a pavement section. PCIer, on the other hand, includes PCI condition categories such as Good, Fair, and Poor, enabling classification-based instruction formats. These datasets link visual features with engineering-based condition ratings.

\begin{figure}[!ht]
\centering
\includegraphics[scale=1.0]{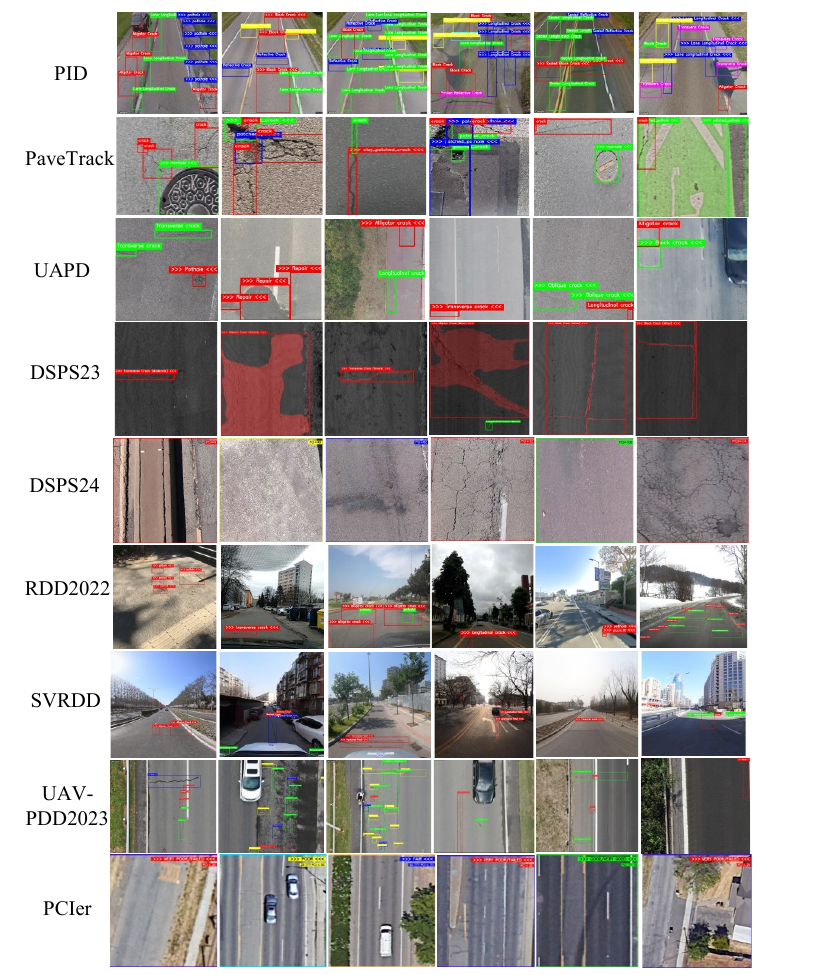}
\caption{Sample annotated images across each of the individual datasets in PaveInstruct.}
\label{sample_images_paveinstruct}
\end{figure}

Although the raw datasets differ in format and task emphasis, together they support a wide set of instruction types across detection, segmentaion, severity interpretation, and PCI estimation. Each dataset offers unique elements, such as patched classes, sealed crack variants, or severity levels, which help capture the complexity of real pavement evaluation. The varied image acquisition methods, including UAVs, smartphones, street-view platforms, and infrastructure-mounted sensors, further ensure that the instructions generated from these sources reflect diverse and realistic field conditions. Figure \ref{sample_images_paveinstruct} shows some of the images for different datasets used in PaveInstruct.

\begin{table}[!h]
    \centering
    \caption{Summary of Pavement Distress Datasets}
    \label{tab:raw_datasets}
    \begin{adjustbox}{max width=\textwidth}
    \begin{tabular}{@{}l c l l l l@{}}
        \toprule
        \textbf{Dataset} & \textbf{\# Images} & \textbf{Distress Types} & \textbf{Data Source} & \textbf{Geographic Regions} & \textbf{Image Perspectives} \\
        \midrule
        PID \cite{PID} & 7,237 & 
        \begin{tabular}[t]{@{}l@{}}
            Reflective cracks, \\
            Transverse cracks, \\
            Block cracks, \\
            Longitudinal cracks, \\
            Alligator cracks, PCC, \\
            Potholes
        \end{tabular} & 
        \begin{tabular}[t]{@{}l@{}}
            Google Street View API \\
        \end{tabular} & 
        \begin{tabular}[t]{@{}l@{}}
            United States
        \end{tabular} & 
        \begin{tabular}[t]{@{}l@{}}
            Wide-view and \\
            top-down aerial view
        \end{tabular} \\
        \midrule
        PaveTrack (US) \cite{PaveTrack} & 5,987 & 
        \begin{tabular}[t]{@{}l@{}}
            Crack, Pothole, \\
            Patched crack, \\
            Patched pothole, \\
            Clay-patched crack, \\
            Manhole
        \end{tabular} & 
        \begin{tabular}[t]{@{}l@{}}
            Mobile vehicle
        \end{tabular} & 
        \begin{tabular}[t]{@{}l@{}}
            California (United States) \\
        \end{tabular} & 
        \begin{tabular}[t]{@{}l@{}}
            Top-down aerial view
        \end{tabular} \\
        \midrule
        UAPD \cite{UAPD} & 3,151 & 
        \begin{tabular}[t]{@{}l@{}}
            Transverse cracks, \\
            Longitudinal cracks, \\
            Alligator cracks, \\
            Oblique cracks, \\
            Potholes, Repairs
        \end{tabular} & 
        \begin{tabular}[t]{@{}l@{}}
            UAV-based
        \end{tabular} & 
        \begin{tabular}[t]{@{}l@{}}
            Nanjing (China)
        \end{tabular} & 
        \begin{tabular}[t]{@{}l@{}}
            Top-down aerial view
        \end{tabular} \\
        \midrule
        DSPS23 \cite{DSPS} & 108 & 
        \begin{tabular}[t]{@{}l@{}}
            Transverse cracks,\\
            Longitudinal cracks,\\
            Alligator cracks,\\
            Block cracks,\\
            Manhole,\\
            Patching
        \end{tabular} & 
        \begin{tabular}[t]{@{}l@{}}
            Synthetic dataset
        \end{tabular} & 
        \begin{tabular}[t]{@{}l@{}}
           ---
        \end{tabular} & 
        \begin{tabular}[t]{@{}l@{}}
            Top-down view
        \end{tabular} \\
        \midrule
        DSPS24 \cite{DSPS} & 7,000 & 
        \begin{tabular}[t]{@{}l@{}}
            Numerical PCI values\\ 
            (0--100)
        \end{tabular} & 
        \begin{tabular}[t]{@{}l@{}}
            Infrastructure-mounted \\
            sensors
        \end{tabular} & 
        \begin{tabular}[t]{@{}l@{}}
           Jefferson City -- Missouri, Peoria, \\ Washington -- Illinois (United States)
        \end{tabular} & 
        \begin{tabular}[t]{@{}l@{}}
            Top-down view
        \end{tabular} \\
        \midrule
        RDD2022 \cite{RDD2022} & 47,420 & 
        \begin{tabular}[t]{@{}l@{}}
            Longitudinal cracks, \\
            Transverse cracks, \\
            Alligator cracks, \\
            Potholes, \\
            Other corruption
        \end{tabular} & 
        \begin{tabular}[t]{@{}l@{}}
            Mixed: Smartphone, \\
            Drone, Google Street View
        \end{tabular} & 
        \begin{tabular}[t]{@{}l@{}}
            Japan, India, Czech Republic, \\
            Norway, United States, China
        \end{tabular} & 
        \begin{tabular}[t]{@{}l@{}}
            Wide, extra-wide, and \\
            top-down views
        \end{tabular} \\
        \midrule
        SVRDD \cite{SVRDD} & 8,000 & 
        \begin{tabular}[t]{@{}l@{}}
            Longitudinal cracks, \\
            Transverse cracks, \\
            Alligator cracks, \\
            Longitudinal patches, \\
            Transverse patches, \\
            Manhole covers
        \end{tabular} & 
        \begin{tabular}[t]{@{}l@{}}
            Baidu Street View \\
        \end{tabular} & 
        \begin{tabular}[t]{@{}l@{}}
            Beijing (China)
        \end{tabular} & 
        \begin{tabular}[t]{@{}l@{}}
            Street view \\
            (front-facing perspective)
        \end{tabular} \\
        \midrule
        UAV-PDD2023 \cite{UAV_PDD23} & 2,440 & 
        \begin{tabular}[t]{@{}l@{}}
            Longitudinal cracks, \\
            Transverse cracks, \\
            Alligator cracks, \\
            Oblique cracks, \\
            Patching, Potholes
        \end{tabular} & 
        \begin{tabular}[t]{@{}l@{}}
            UAV aerial capture \\
        \end{tabular} & 
        \begin{tabular}[t]{@{}l@{}}
             Tianjin (China)
        \end{tabular} & 
        \begin{tabular}[t]{@{}l@{}}
            Top-down aerial view \\
        \end{tabular} \\
        \midrule
        PCIer \cite{PCIer} & 480 & 
        \begin{tabular}[t]{@{}l@{}}
            PCI condition classes: \\
             Good (70--100),\\
             Fair (50--69),\\
             Poor (25--49),\\
             Very Poor/Failed (0--24)
        \end{tabular} & 
        \begin{tabular}[t]{@{}l@{}}
            Google Earth \\
        \end{tabular} & 
        \begin{tabular}[t]{@{}l@{}}
             California (United States)
        \end{tabular} & 
        \begin{tabular}[t]{@{}l@{}}
            Aerial view
        \end{tabular} \\
        \midrule
        Other sources & 7,590 & 
        \begin{tabular}[t]{@{}l@{}}
            Longitudinal cracks, \\
            Transverse cracks, \\
            Alligator cracks, \\
            Potholes, Ruts, \\
            Edge cracking, \\
            Patching
        \end{tabular} & 
        \begin{tabular}[t]{@{}l@{}}     
            Online source \\
            (Roboflow Universe)
        \end{tabular} & 
        \begin{tabular}[t]{@{}l@{}}
                ---
        \end{tabular} & 
        \begin{tabular}[t]{@{}l@{}}
            Top-down and perspective\\
            aerial views
        \end{tabular} \\
        \bottomrule
    \end{tabular}
    \end{adjustbox}
\end{table}

\subsection{Instruction Generation Pipeline}
Our instruction-generation process is inspired by the design principles of the LLaVA-Instruct-150K dataset \cite{liu2023llavainstruct150k}, which we tailor to the pavement infrastructure domain. 
\subsubsection*{Task Taxonomy and Generation Framework}
\begin{figure}[!ht]
\centering
\includegraphics[scale=1.1]{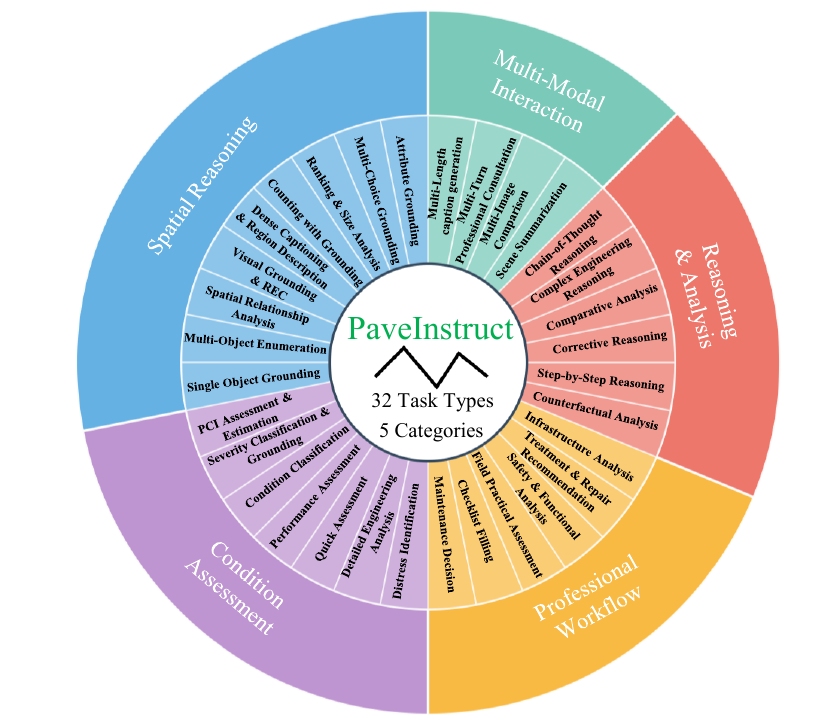}
\caption{Overview of the PaveInstruct task taxonomy}
\label{paveinstruct_categories}
\end{figure}

The construction of our instruction-following dataset is grounded in a comprehensive taxonomy of task types that reflects the diverse cognitive, spatial, and professional competencies required for pavement understanding. These tasks are organized into five broad categories: \textit{Spatial Reasoning Tasks}, \textit{Condition Assessment Tasks}, \textit{Professional Workflow Tasks}, \textit{Reasoning and Analysis Tasks}, and \textit{Multi-Modal Interaction Tasks}. This taxonomy is designed to capture the full spectrum of interactions between visual inputs, spatial cues, engineering-level judgments, and professional decision-making processes essential for comprehensive pavement infrastructure management. Figure \ref{paveinstruct_categories} shows a summary of all the different task categories and their corresponding sub-tasks. 

\vspace{1mm}
\noindent\textbf{Spatial Reasoning Tasks:} These tasks elicit complex spatial understanding and localization capabilities from the model, encompassing both precise coordinate-based reasoning and complex spatial relationship analysis. They are critical for training models to perceive, ground, and reason about pavement distresses within diverse visual contexts. Below are some of the sub-tasks under the spatial reasoning. 

\begin{itemize}
    \item \textit{Single Object Grounding:} This task requires the model to precisely identify and localize individual pavement distresses through natural language queries, providing exact bounding box coordinates for specific distress instances. E.g.\ ``find the largest pothole in the wheel path.''
    
    \item \textit{Multi-Object Enumeration:} This capability involves systematic identification and spatial enumeration of all instances within specific distress categories, including comprehensive listing with coordinate verification and spatial distribution analysis. E.g:\ ``list all alligator cracks with their coordinates" or \ ``enumerate all potholes from largest to smallest with bounding boxes."
    
    \item \textit{Spatial Relationship Analysis:} This task focuses on complex geometric reasoning about relationships between pavement elements, including proximity analysis, intersection detection, and relative positioning assessment of distress patterns. E.g: \ ``which crack intersects with the patched area?"
    
    \item \textit{Visual Grounding and Referring Expression Comprehension:} This capability enables localization of pavement regions through complex natural language descriptions that integrate spatial attributes, distress characteristics, and contextual positioning information. E.g: \ ``the spalled area adjacent to the manhole cover."
    
    \item \textit{Dense Captioning and Region Description:} This task involves generating detailed, spatially-referenced technical descriptions for specific pavement regions, integrating coordinate-based localization with comprehensive distress characterization. E.g: \ ``Region [245,156,678,344]: Medium-severity alligator crack with interconnected pattern showing edge spalling."
    
    \item \textit{Counting with Grounding:} This capability combines systematic quantitative enumeration with spatial coordinate verification, ensuring accurate distress counting with precise location documentation for validation purposes. E.g: \ ``count all transverse cracks and provide their coordinates".
    
    \item \textit{Ranking and Size Analysis:} This task focuses on comparative spatial assessment of distress instances based on size, area, and severity metrics, requiring quantitative analysis and priority-based ordering capabilities. E.g: \ ``rank all distresses by severity from low to high".
    
    \item \textit{Multi-Choice Grounding:} This capability involves structured spatial reasoning questions with engineering-relevant alternatives, requiring precise visual discrimination and accurate localization among similar distress types. E.g: \ ``Which distress is in the upper-left quadrant? (a) center pothole, (b) right-side crack, (c) top-left alligator crack."
    
    \item \textit{Attribute Grounding:} This task emphasizes detailed analysis of distress-specific visual characteristics and material properties, connecting spatial localization with technical attribute identification and professional assessment criteria. E.g: \ ``assess the patch material quality at coordinates [100,200,300,400]".
\end{itemize}

\vspace{1mm}
\noindent\textbf{Condition Assessment Tasks:} This category emphasizes systematic condition evaluation, diagnostic inference, and engineering-based assessment consistent with professional pavement management standards. These tasks are essential for replicating real-world pavement inspection and evaluation workflows.

\begin{itemize}
    \item \textit{PCI Assessment and Estimation:} This task involves systematic estimation of Pavement Condition Index values following ASTM D6433 methodology, including comprehensive reasoning chains that connect observed distresses to quantitative condition ratings. E.g: \ ``provide a PCI rating with step-by-step ASTM D6433 reasoning."
    
    \item \textit{Severity Classification and Grounding:} This capability focuses on detailed classification of individual distress severity levels using professional criteria, requiring evidence-based justification through specific visual indicators and engineering assessment standards. E.g: \ ``classify this alligator crack as Low/Medium/High severity and justify using ASTM criteria" or \ ``assess the severity level of the pothole at coordinates [150,250,300,350]."
    
    \item \textit{Condition Classification:} This task involves systematic assignment of overall pavement condition categories based on comprehensive distress analysis, incorporating structural integrity assessment and functional performance evaluation. E.g: \ ``classify this pavement as Excellent/Good
    /Fair/Poor/Failed based on visible distresses" or \ ``determine the overall condition rating and provide supporting evidence."
    
    \item \textit{Performance Assessment:} This capability emphasizes evaluation of current pavement functional capacity and prediction of performance degradation patterns based on observable distress characteristics and structural condition indicators. E.g: \ ``assess how these distresses impact ride quality and vehicle operations".
    
    \item \textit{Quick Assessment:} This task focuses on streamlined evaluation methodologies designed for immediate field decision-making, providing rapid but accurate condition classifications for operational efficiency. E.g: \ ``immediate repair needed? (Yes/No)".
    
    \item \textit{Detailed Engineering Analysis:} This capability involves comprehensive technical evaluation that integrates multiple distress interactions, failure mechanism analysis, and systematic engineering assessment methodologies for complex pavement conditions. E.g: \ ``analyze the interaction between fatigue cracking and environmental deterioration".
    
    \item \textit{Distress Identification:} This task emphasizes systematic recognition and professional classification of specific pavement failure modes, requiring accurate application of technical nomenclature and diagnostic criteria. E.g: \ ``identify and classify all visible distress types using ASTM terminology".
\end{itemize}

\vspace{1mm}
\noindent\textbf{Professional Workflow Tasks:} These tasks replicate professional pavement management workflows, incorporating industry-standard practices, documentation requirements, and decision-making protocols used in real-world infrastructure management.

\begin{itemize}
    \item \textit{Infrastructure Analysis:} This capability involves comprehensive assessment of pavement infrastructure elements and their interaction with overall pavement condition, including evaluation of repair effectiveness and asset management implications. E.g: \ ``evaluate the performance of existing patches".
    
    \item \textit{Treatment and Repair Recommendation:} This task focuses on development of specific maintenance and rehabilitation strategies based on observed conditions, incorporating professional standards, cost-effectiveness analysis, and treatment prioritization methodologies. E.g: \ ``recommend specific repair treatments for each identified distress".
    
    \item \textit{Safety and Functional Analysis:} This capability emphasizes evaluation of distress impacts on vehicle operations, traffic safety, and functional capacity, including risk assessment and operational mitigation strategy development. E.g: \ ``assess tire damage risk from these potholes".
    
    \item \textit{Field Practical Assessment:} This task involves simulation of actual pavement inspection scenarios under real-world constraints, including equipment limitations, environmental factors, and practical decision-making requirements. E.g: \ ``conduct inspection under time constraints".
    
    \item \textit{Checklist Filling:} This capability focuses on systematic completion of standardized assessment documentation and regulatory compliance protocols, ensuring proper data collection and professional reporting standards. E.g: \ ``complete ASTM D6433 survey form for this section".
    
    \item \textit{Maintenance Decision:} This task emphasizes strategic decision-making processes involving resource allocation, timing optimization, and comprehensive pavement management program development based on condition assessment findings. E.g: \ ``prioritize repair schedule for budget allocation".
\end{itemize}

\vspace{1mm}
\noindent\textbf{Reasoning and Analysis Tasks:} This category focuses on demonstrating systematic analytical thinking and professional reasoning processes, training models to exhibit transparent decision-making methodologies consistent with engineering practice.

\begin{itemize}
    \item \textit{Chain-of-Thought Reasoning:} This capability involves systematic demonstration of professional assessment methodology, showing explicit progression from initial observation through analytical stages to final recommendations with transparent reasoning chains. E.g: \ ``walk through your assessment process step-by-step from initial observation to final PCI calculation".
    
    \item \textit{Complex Engineering Reasoning:} This task focuses on detailed analytical integration of multiple distress factors, material considerations, and structural implications to produce comprehensive condition assessments and strategic recommendations. E.g: \ ``analyze how traffic loading, environmental factors, and material aging interact to produce this distress pattern".
    
    \item \textit{Comparative Analysis:} This capability emphasizes systematic comparison methodologies for multiple distresses or pavement sections, including relative assessment, priority ranking, and resource allocation decision-making with quantitative justification. E.g: \ ``compare repair urgency between the pothole and alligator crack".
    
    \item \textit{Corrective Reasoning:} This task involves identification and correction of assessment errors and professional misconceptions, promoting quality assurance and accurate judgment development through error analysis and educational guidance. E.g: \ ``this crack was misclassified as high severity - explain why it's actually medium severity".
    
    \item \textit{Step-by-Step Reasoning:} This capability focuses on methodical demonstration of professional assessment protocols, showing systematic progression through evaluation stages with explicit justification for each analytical decision point. E.g: \ ``demonstrate the systematic procedure for PCI calculation with explicit steps".
    
    \item \textit{Counterfactual Analysis:} This task emphasizes analytical examination of alternative scenarios and hypothetical conditions, requiring predictive reasoning about potential outcomes and alternative assessment interpretations under different circumstances. E.g: \ ``what would happen if this alligator crack were left untreated for two years?".
\end{itemize}

\vspace{1mm}
\noindent\textbf{Multi-Modal Interaction Tasks:} These tasks encompass complex interaction modalities that integrate visual analysis with diverse response formats, conversation patterns, and professional communication requirements.

\begin{itemize}
    \item \textit{Multi-Length Caption Generation:} This capability involves production of technical descriptions at varying complexity levels tailored to different professional communication contexts, from brief field documentation to comprehensive inspection reporting requirements. E.g: brief field note \ ``3 potholes, medium severity"
    
    \item \textit{Multi-Turn Professional Consultation:} This task focuses on extended technical dialogues that simulate real engineering consultations, maintaining progressive complexity development and technical coherence across multiple conversational exchanges. E.g: progressive dialogue from \ ``what distresses do you see?" to \ ``which requires priority?" to \ ``what treatment strategy?" to \ ``what's the implementation timeline?"
    
    \item \textit{Multi-Image Comparison:} This capability emphasizes systematic comparative analysis across multiple pavement images, requiring integration of visual evidence from diverse sources and comprehensive condition assessment across different infrastructure sections. E.g: \ ``compare the deterioration levels between these three pavement sections".
    
    \item \textit{Scene Summarization:} This task involves comprehensive synthesis of overall pavement condition based on multiple distress distributions and infrastructure elements, providing executive-level summaries suitable for management decision-making and strategic planning. E.g: \ ``summarize critical maintenance needs across this pavement section."
\end{itemize}

\vspace{1mm}
\subsubsection{Multi-Source Dataset Integration Pipeline}

\begin{figure}[!ht]
\centering
\includegraphics[scale=0.42]{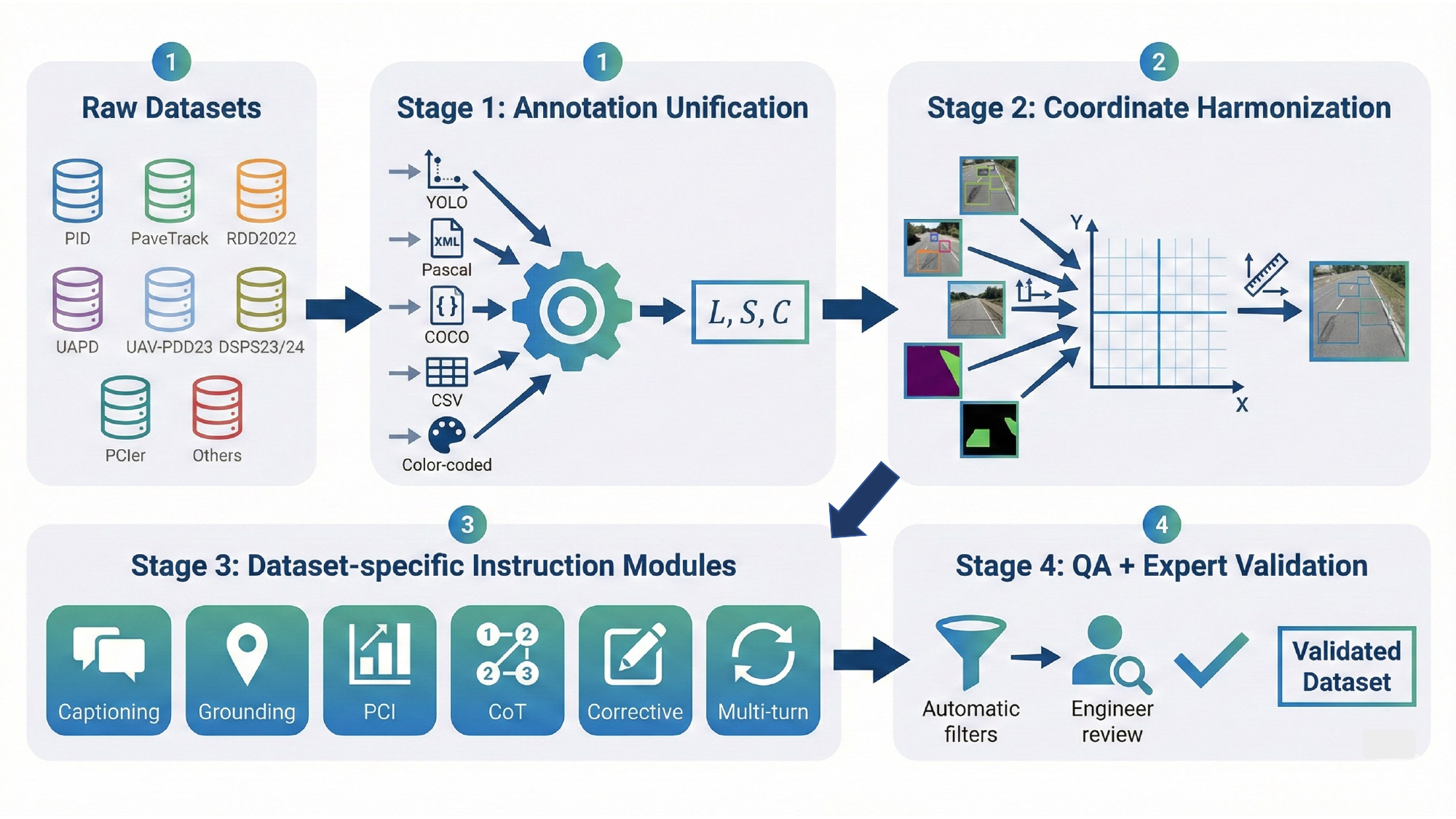}
\caption{PaveInstruct Dataset Creation Pipeline}
\label{paveinstruct_framework}
\end{figure}

The creation of PaveInstruct requires systematic integration of annotations across heterogeneous data sources, each employing distinct annotation schemas optimized for specific computer vision tasks. This integration challenge stems from the diverse origins and intended applications of the source datasets, which collectively span object detection, segmentation, condition assessment, and infrastructure management domains. We address this complexity through a structured four-stage pipeline that preserves annotation semantics while establishing consistency. Figure \ref{paveinstruct_framework} shows the overall framework for PaveInstruct creation. 

\vspace{2mm}
\noindent\textbf{Stage 1: Annotation Format Unification}

The initial stage addresses the fundamental heterogeneity of annotation formats across our nine source datasets. These datasets employ four distinct annotation schemas: YOLO normalized coordinates, Pascal VOC XML structures, color-coded classification systems, and CSV-based condition ratings. Table \ref{tab:annotation_formats} shows the different annotation formats for the datasets. Each format reflects the computational requirements and domain conventions of its respective research community, creating incompatibilities that must be resolved before instruction generation.

\begin{table}[h]
    \centering
    \caption{Dataset Annotation Formats and Task Types}
    \label{tab:annotation_formats}
    \begin{tabular}{@{}l l l@{}}
        \toprule
        \textbf{Dataset} & \textbf{Annotation Format} & \textbf{Task Type} \\
        \midrule
        PID \cite{PID} & YOLO & Distress Detection \\
        \midrule
        PaveTrack (US) \cite{PaveTrack} & YOLO & 
        \begin{tabular}[t]{@{}l@{}}
            Distress Detection
        \end{tabular} \\
        \midrule
        UAPD \cite{UAPD} & Pascal VOC XML & Distress Detection \\
        \midrule
        DSPS23 \cite{DSPS} & COCO & Severity Estimation \\
        \midrule
        DSPS24 \cite{DSPS} & CSV & PCI Assessment \\
        \midrule
        RDD2022 \cite{RDD2022} & YOLO & Distress Detection \\
        \midrule
        SVRDD \cite{SVRDD} & YOLO & Distress Detection \\
        \midrule
        UAV-PDD2023 \cite{UAV_PDD23} & Pascal VOC XML & Distress Detection \\
        \midrule
        PCIer \cite{PCIer} & Color-coded Folders & 
        \begin{tabular}[t]{@{}l@{}}
            PCI Classification
        \end{tabular} \\
        \bottomrule
    \end{tabular}
\end{table}

YOLO format datasets, including RDD2022, PaveTrack, and SVRDD, encode bounding boxes as normalized center coordinates with relative width and height:

\begin{equation}
    \mathbf{b}_{yolo} = (x_c, y_c, w, h) \in [0,1]^4
\end{equation}

where $x_c, y_c$ represent normalized center coordinates, and $w, h$ denote relative width and height respectively. Pascal VOC datasets, such as UAPD and UAV-PDD2023, specify absolute pixel coordinates as corner points:

\begin{equation}
    \mathbf{b}_{voc} = (x_{min}, y_{min}, x_{max}, y_{max}) \in \mathbb{R}^4
\end{equation}

where $(x_{min}, y_{min})$ and $(x_{max}, y_{max})$ represent top-left and bottom-right corner coordinates respectively. Color-coded datasets like PCIer represent condition classifications through chromatic encoding:

\begin{equation}
\begin{split}
    \mathbf{c} = \{Green, Blue, Yellow, Red\} \\
    \mapsto \{Good, Fair, Poor, Failed\}
\end{split}
\end{equation}

CSV-based datasets such as DSPS24 provide direct numeric condition indices:

\begin{equation}
    s \in [0,100]
\end{equation}

where $s$ represents the pavement condition index score.

The unification process transforms these heterogeneous representations into a standardized annotation schema $\mathcal{A}_{unified}$ that preserves semantic content while enabling consistent computational processing. We define the transformation function:

\begin{equation}
    \mathcal{A}_{unified} = \mathcal{U}(\mathcal{A}_{yolo}, \mathcal{A}_{pascal}, \mathcal{A}_{color}, \mathcal{A}_{csv})
\end{equation}

where $\mathcal{U}$ represents the unification operator that maps diverse annotation formats to a common representation space, and $\mathcal{A}_{yolo}, \mathcal{A}_{pascal}, \mathcal{A}_{color}, \mathcal{A}_{csv}$ denote annotations from YOLO, Pascal VOC, color-coded, and CSV formats respectively. This operator applies format-specific conversion functions while maintaining annotation integrity:

\begin{equation}
    \mathcal{U} = \phi_{yolo} \circ \phi_{pascal} \circ \phi_{color} \circ \phi_{csv}
\end{equation}

where $\phi_{yolo}, \phi_{pascal}, \phi_{color}, \phi_{csv}$ represent format-specific transformation functions. The YOLO-to-unified transformation $\phi_{yolo}$ converts normalized coordinates to absolute pixel coordinates:

\begin{equation}
\begin{split}
    \phi_{yolo}(\mathbf{b}_{yolo}) = &(x_c \cdot W - w \cdot W/2, y_c \cdot H - h \cdot H/2, \\
    &\phantom{(}x_c \cdot W + w \cdot W/2, y_c \cdot H + h \cdot H/2)
\end{split}
\end{equation}

where $(W, H)$ denotes image dimensions in pixels. The Pascal VOC transformation $\phi_{pascal}$ preserves absolute coordinates while ensuring consistent indexing conventions. Color-coded transformations $\phi_{color}$ map chromatic classifications to standardized condition categories, while CSV transformations $\phi_{csv}$ integrate numeric indices with spatial metadata when available.

The resulting unified annotation schema maintains three core components: spatial localization data $\mathbf{L} \in \mathbb{R}^4$ for distress positioning, semantic classification $\mathbf{S}$ for distress type identification, and condition assessment $\mathbf{C}$ for severity or PCI rating. This standardization enables consistent instruction generation across all source datasets while preserving the semantic richness inherent in each original annotation format. The unified representation supports both spatial reasoning tasks that require precise localization and assessment tasks that emphasize condition evaluation and maintenance decision-making.

\vspace{2mm}
\noindent\textbf{Stage 2: Coordinate System Harmonization}

The second stage addresses geometric inconsistencies in bounding box coordinates arising from diverse image resolutions and aspect ratios across source datasets. Raw bounding box annotations are tied to original image dimensions, creating coordinate spaces that vary significantly between datasets. This variability impedes spatial reasoning tasks that require consistent geometric relationships for distress localization.

Let $\mathbf{b}_{orig} \in \mathbb{R}^4$ denote original bounding box coordinates and $(H_{orig}, W_{orig})$ represent original image dimensions. We implement a bounding box normalization process that transforms coordinates to a standardized reference frame while preserving spatial relationships. The coordinate transformation function $\mathcal{T}_{bbox}$ maps original bounding boxes to normalized coordinate space:

\begin{equation}
    \mathbf{b}_{norm} = \mathcal{T}_{bbox}(\mathbf{b}_{orig}, H_{orig}, W_{orig}, H_{target}, W_{target})
\end{equation}

where $\mathbf{b}_{norm} \in \mathbb{R}^4$ represents normalized bounding box coordinates and $(H_{target}, W_{target})$ denotes target image dimensions. The transformation employs scaling factors:

\begin{equation}
    s_x = \frac{W_{target}}{W_{orig}}, \quad s_y = \frac{H_{target}}{H_{orig}}
\end{equation}

to ensure proportional bounding box adjustment across coordinate systems.

The harmonization process computes spatial relationship matrices that capture geometric dependencies between bounding box pairs:

\begin{equation}
    \mathbf{R}_{spatial} \in \mathbb{R}^{n \times n}
\end{equation}

where $n$ represents the number of distress instances and $\mathbf{R}_{spatial}$ encodes proximity measures, overlap ratios, and directional relationships. This unified coordinate system enables consistent spatial reasoning for instruction generation tasks requiring precise distress localization and cross-dataset geometric analysis.

\vspace{2mm}
\noindent\textbf{Stage 3: Task-Specific Instruction Generation}

The third stage implements domain-constrained synthesis to generate instruction-response pairs based on the established task taxonomy and different dataset-specific instruction modules. This process utilizes a large language model as controlled generation engine within systematic engineering frameworks, ensuring technical accuracy while maintaining instructional diversity across both spatial reasoning and assessment task categories.

We define the task-specific generation function $\Gamma_{\mathcal{T}}$ that produces instruction-response pairs for task type $\mathcal{T}$:

\begin{equation}
    (\mathbf{q}_i, \mathbf{r}_i) = \Gamma_{\mathcal{T}}(\mathcal{V}_i, \mathcal{A}_{unified}, \Psi_{\mathcal{T}}, \mathcal{L}_j)
\end{equation}

where $\mathbf{q}_i$ represents the generated instruction, $\mathbf{r}_i$ denotes the corresponding response, $\Psi_{\mathcal{T}}$ encodes task-specific system prompts, and $\mathcal{L}_j \in \{short, medium, long\}$ specifies instruction length variants.

\vspace{1mm}
\noindent{\textbf{Prompt Structure and Templates.}} Each task category employs specialized system prompts that inject domain expertise and constrain generation within professional engineering workflows. The prompt structure $\Psi_{\mathcal{T}}$ systematically integrates three components:

\begin{equation}
    \Psi_{\mathcal{T}} = \Omega_{domain} \oplus \Sigma_{ASTM} \oplus \Theta_{task}
\end{equation}

where $\Omega_{domain}$ enforces pavement engineering terminology, $\Sigma_{ASTM}$ injects ASTM D6433 compliance requirements, and $\Theta_{task}$ provides task-specific generation constraints.

To apply this structure in practice, we define six broad template families: captioning, chain-of-thought, grounding, PCI-specific, corrective, and multi-turn. Captioning templates generate short, medium, or long descriptions of visual content. Chain-of-thought prompts elicit step-by-step reasoning, especially for PCI estimation and severity analysis. Grounding templates handle localization, spatial comparison, object ranking, and region descriptions. PCI-specific prompts include condition estimation, justification, checklist filling, and treatment suggestions. Corrective prompts simulate user mistakes and trigger expert clarifications. Multi-turn templates model progressive engineering conversations from initial inspection to maintenance planning. These template families ensure that instruction generation is task-aligned, diverse, and reflective of real-world pavement evaluation workflows. Each template family supports multiple complexity levels, described next.

\vspace{1mm}
\noindent\textbf{Variable Length Generation.} The generation process produces instruction variants across three complexity levels to ensure comprehensive model training:

\begin{equation}
    \mathcal{I}_{\mathcal{L}} = \{\Gamma_{\mathcal{T}}(\mathcal{V}_i, \mathcal{A}_{unified}, \Psi_{\mathcal{T}}, \mathcal{L}_j)\}_{\mathcal{L}_j}
\end{equation}
where $\mathcal{L}_j \in \{short, medium, long\}$.

where short instructions focus on direct identification, medium instructions incorporate contextual reasoning, and long instructions demand comprehensive technical analysis with detailed justifications.

\vspace{1mm}
\noindent\textbf{Multi-turn Conversation Creation.} 
As one of the six template families, multi-turn prompts model engineering conversations across multiple dialogue turns. Progressive technical dialogues simulate real-world engineering consultations through structured conversation patterns:

\begin{equation}
    \mathcal{C}_k = \Gamma_{multi}(\mathcal{V}_i, \mathcal{A}_{unified}, \{\Psi_t\}_{t=1}^{T})
\end{equation}

where $\mathcal{C}_k$ represents conversation $k$ with $T$ turns, and $\{\Psi_t\}_{t=1}^{T}$ denotes turn-specific prompts that progressively increase technical depth from initial observation through detailed analysis to maintenance recommendations.

\vspace{2mm}
\noindent\textbf{Stage 4: Quality Assurance and Validation}

To ensure the accuracy and engineering relevance of the generated instruction-response pairs, we implement a multi-stage quality assurance process that combines automated validation with expert human review. Each sample is first checked for structural integrity, including the presence of required fields such as bounding boxes, severity labels, or PCI values, depending on task type. Domain-specific consistency checks are then applied to verify that the terminology and reasoning used in responses align with ASTM D6433 standards and pavement engineering conventions. For instance, PCI predictions are validated to fall within the permissible range [0, 100] and to correspond logically with described distresses. Beyond these automated checks, we conduct targeted human validation by licensed pavement engineers and domain experts, who review a diverse subset of samples for technical correctness, language clarity, and alignment with real-world inspection practices. This expert review stage is critical for identifying subtle inconsistencies, hallucinated reasoning, or misclassifications that automated filters may miss. These layered validation procedures ensure that PaveInstruct provides high-quality, instruction-tuned supervision that reflects professional standards in pavement evaluation and maintenance planning.









\subsubsection{PaveInstruct Dataset Statistics and Analysis}

\begin{figure}[!ht]
\centering
\includegraphics[scale=1.0]{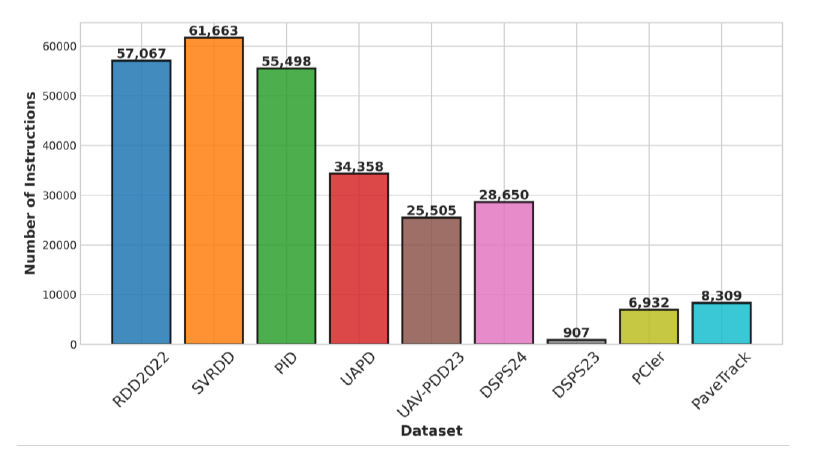}
\caption{Instruction count per source dataset.}
\label{fig:instruction_distribution}
\end{figure}

The final PaveInstruct dataset consists of \textbf{278,889} instruction–response pairs derived from nine core datasets, covering diverse distress types, image perspectives, and condition annotations. Figure~\ref{fig:instruction_distribution} shows the distribution of instruction counts across the source datasets. SVRDD, RDD2022, and PID collectively contribute the largest share of instructions, reflecting their large image pools and high distress density. In contrast, DSPS23 and PCIer contain fewer samples due to limited original annotations or narrower task scopes.

\begin{figure}[H]
\centering
\includegraphics[scale=1.0]{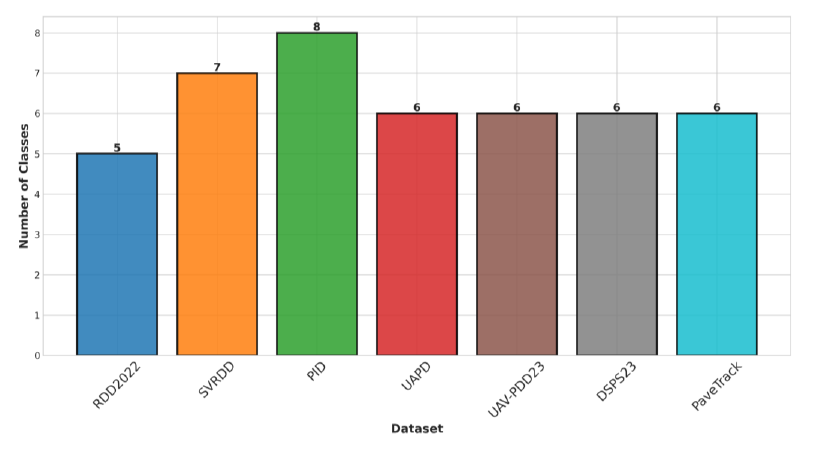}
\caption{Number of distress classes per dataset.}
\label{fig:class_counts}
\end{figure}

Across datasets, the number of annotated distress classes varies, as shown in Figure~\ref{fig:class_counts}. PID has the highest number of distinct distress types (8), followed closely by SVRDD and DSPS23, each with 6–7 categories. Notably, DSPS23’s original fine-grained labels for severity levels have been consolidated into core distress classes, such as alligator, block, longitudinal, and transverse cracks. In contrast, RDD2022 contains only 5 distress types, indicating a more focused annotation scope. 

\begin{figure}[H]
\centering
\includegraphics[scale=1.1
]{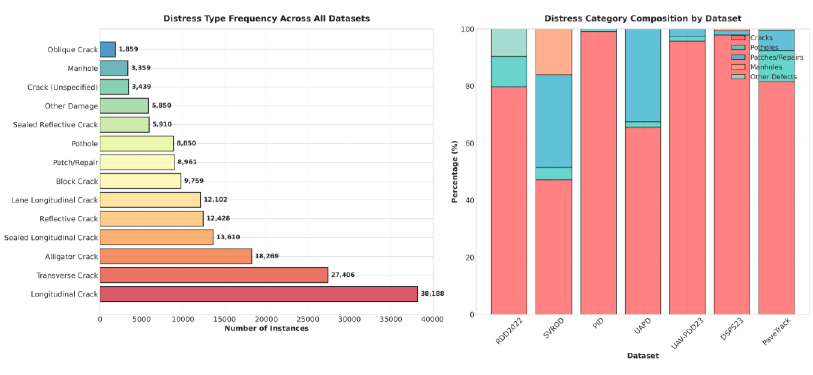}
\caption{(Left) Global distress type frequencies. (Right) Percentage distribution of distress categories by dataset.}
\label{fig:distress_statistics}
\end{figure}

Figure~\ref{fig:distress_statistics} shows the distribution of distress types in PaveInstruct. The left plot confirms that longitudinal, transverse, and alligator cracks are the most common, with longitudinal cracks appearing over 38,000 times. Less frequent distress types include oblique cracks, manholes, and reflective variants. The right plot shows that most datasets such as RDD2022, PID, and UAV-PDD23 are dominated by crack-related annotations. In contrast, SVRDD and DSPS23 contain a more balanced mix, including patches, potholes, manholes, and surface defects. This label diversity enables both crack-focused and general pavement reasoning tasks.

\begin{figure}[H]
\centering
\includegraphics[scale=0.75]{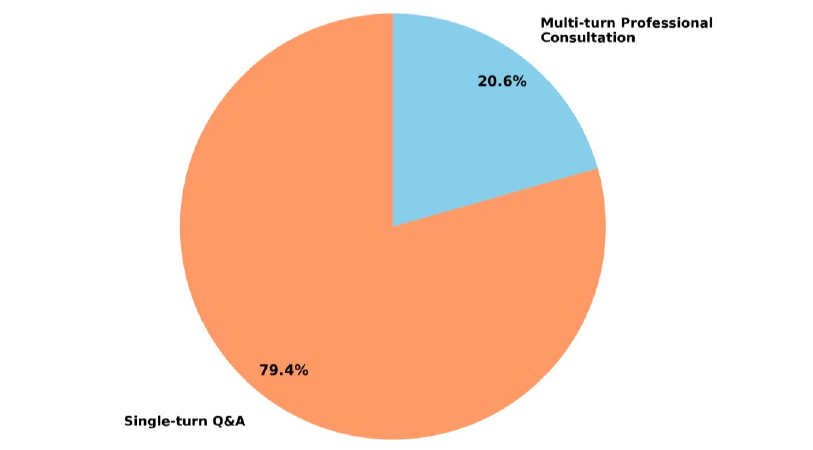}
\caption{Distribution of single-turn vs. multi-turn conversations.}
\label{fig:conversation_types}
\end{figure}

Instruction-response pairs in PaveInstruct span both single-turn and multi-turn interactions. As shown in Figure~\ref{fig:conversation_types}, 20.6\% of samples are multi-turn professional consultations, simulating real-world inspection dialogues. The remaining 79.4\% are single-turn Q\&A, primarily used for direct spatial or diagnostic queries.

\begin{figure}[!ht]
\centering
\includegraphics[scale=1.0]{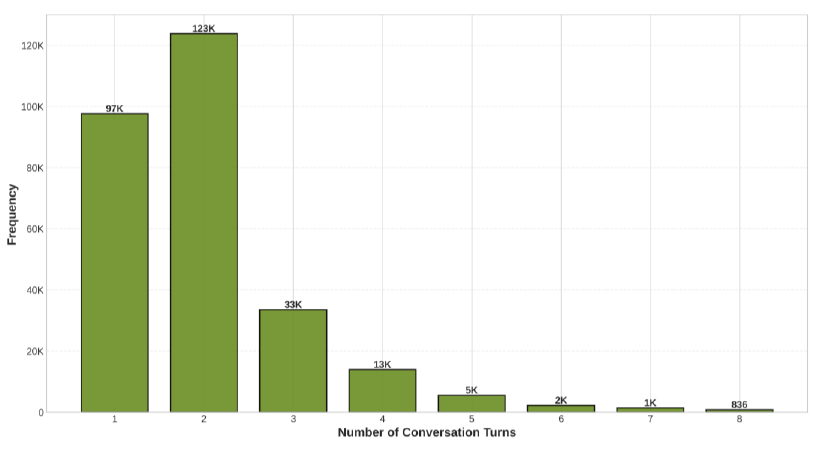}
\caption{Distribution of conversation turns per dialogue.}
\label{fig:conversation_turns}
\end{figure}

Figure~\ref{fig:conversation_turns} further breaks down the number of conversation turns in multi-turn samples. The majority of conversations consist of 2–3 turns, with a long tail of deeper dialogues extending to 7–8 turns, enabling progressive reasoning and contextual follow-up.

\begin{figure}[H]
\centering
\includegraphics[scale=1.0]{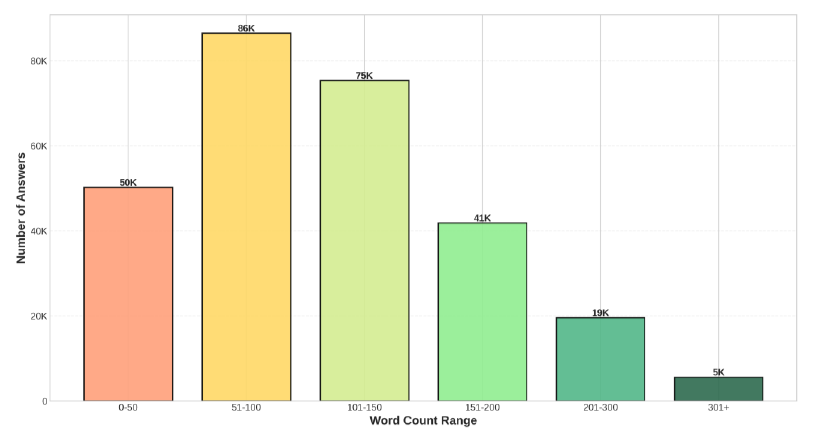}
\caption{Distribution of answer word counts.}
\label{fig:word_count_stats}
\end{figure}

The dataset also supports length variation across answers, reflecting different instruction formats and reasoning depth. Figure~\ref{fig:word_count_stats} shows the distribution of answer lengths. Most responses range between 50–150 words, with a smaller proportion of long-form answers exceeding 200 words, particularly in detailed PCI reasoning or engineering dialogues.

\begin{figure}[H]
\centering
\includegraphics[scale=1.0]{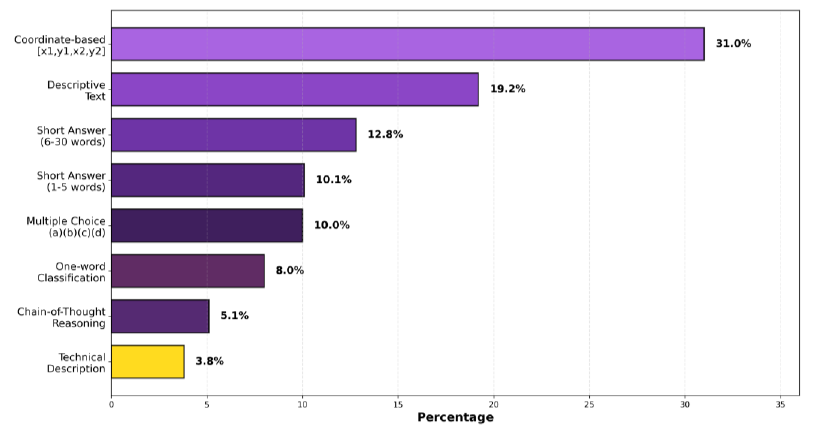}
\caption{Distribution of answer formats in PaveInstruct.}
\label{fig:answer_formats}
\end{figure}

To support diverse training objectives, PaveInstruct instructions are grounded in a range of answer formats. Figure~\ref{fig:answer_formats} illustrates the proportion of coordinate-based outputs (31\%), descriptive text (19.2\%), and other types including short answers, multiple choice, and chain-of-thought reasoning. This variation ensures the dataset supports both visual grounding and structured reasoning tasks.

\begin{figure}[H]
\centering
\includegraphics[scale=1.05]{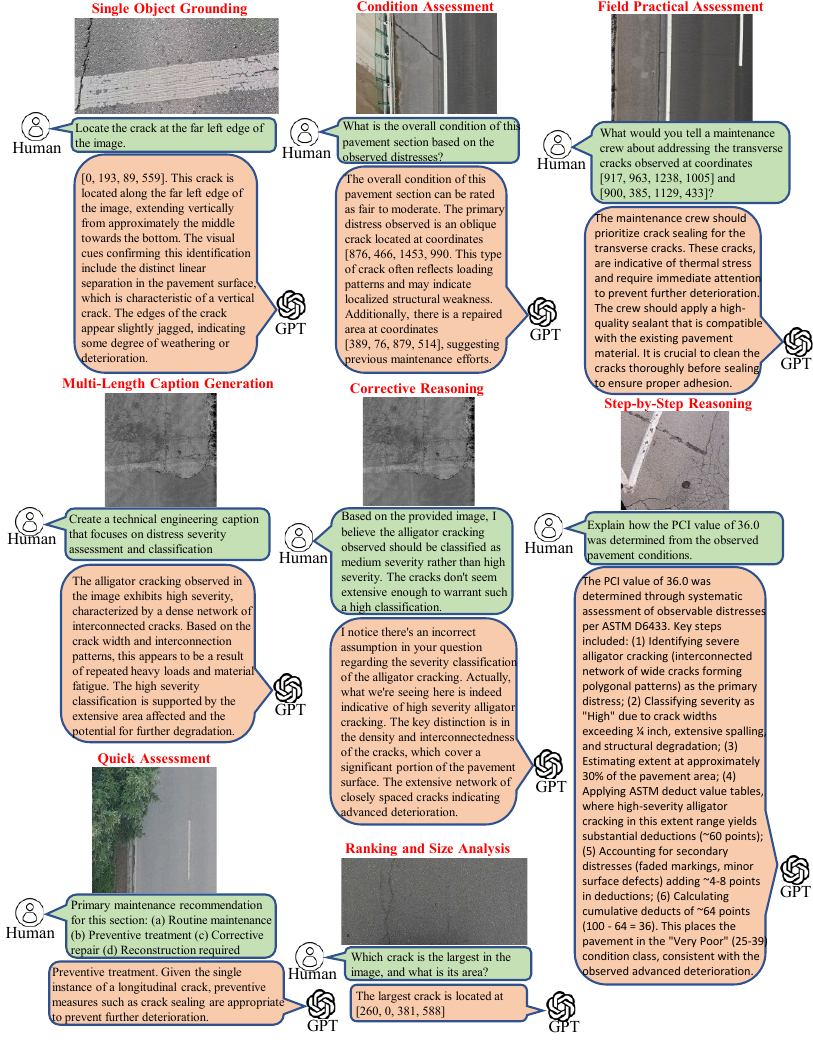}
\caption{Representative instruction-response pairs from PaveInstruct across multiple task families.}
\label{fig:instruction_formats}
\end{figure}

Figure~\ref{fig:instruction_formats} presents representative instruction–response pairs sampled from PaveInstruct across different pavement images and task families. Each example reflects a distinct instruction type, including grounding, condition assessment, maintenance recommendation, captioning, corrective reasoning, PCI estimation, and ranking. While the images vary, all tasks follow a consistent format that links visual input to structured or explanatory responses. Some instructions focus on spatial understanding, such as locating a specific crack, while others require reasoning, such as justifying a PCI score or suggesting appropriate repairs.

\subsection{Problem Structure and Overview}
\label{subsec:problem_overview}
Pavement infrastructure assessment encompasses diverse tasks including visual captioning, visual question answering, object detection, severity classification, PCI estimation, and maintenance recommendation. Traditional approaches require separate specialized models for each task, creating operational challenges for comprehensive evaluation. We address this limitation by developing PaveGPT, a unified vision-language foundation model for pavement assessment. We formulate the problem as follows. 

Given a pavement image $\mathcal{I} \in \mathbb{R}^{H \times W \times 3}$ and a natural language instruction $\mathcal{Q}$, the goal is to generate an appropriate response $\mathcal{R}$. Instructions can request operations such as ``Describe the pavement condition,'' ``Locate all potholes with bounding boxes,'' or ``Estimate the PCI value.'' The response space $\mathcal{R}$ is a union of multiple output modalities:
\begin{equation}
\mathcal{R} = \mathcal{R}_{text} \cup \mathcal{R}_{bbox} \cup \mathcal{R}_{numeric} \cup \mathcal{R}_{class}
\end{equation}
where $\mathcal{R}_{text}$ represents natural language responses, $\mathcal{R}_{bbox} \subset \mathbb{R}^{N \times 4}$ denotes bounding box coordinates, $\mathcal{R}_{numeric} \in \mathbb{R}$ captures continuous values like PCI scores, and $\mathcal{R}_{class}$ represents discrete labels such as severity levels. The core challenge is learning a unified model $f_{\theta}$ that maps instruction-image pairs to appropriate responses:
\begin{equation}
f_{\theta}: (\mathcal{I}, \mathcal{Q}) \rightarrow \mathcal{R}
\end{equation}

PaveGPT builds upon a vision-language architecture consisting of a vision transformer encoder $\mathcal{E}_v$, a large language model backbone $\mathcal{M}_{llm}$, and cross-modal projection layers $\mathcal{P}$. The image is encoded into visual tokens $\mathbf{V} = \mathcal{E}_v(\mathcal{I}) \in \mathbb{R}^{L_v \times d}$ and aligned to the language model's embedding space through projection $\mathbf{V}' = \mathcal{P}(\mathbf{V})$. The instruction is tokenized as $\mathbf{Q} = [q_1, q_2, ..., q_{L_q}]$. PaveGPT generates responses autoregressively:
\begin{equation}
\mathcal{R} = \mathcal{M}_{llm}([\mathbf{V}', \mathbf{Q}]; \theta)
\end{equation}
This architecture treats all outputs as sequences, enabling unified handling of diverse formats. Bounding box coordinates and PCI scores are generated as text tokens representing numerical values.

We train PaveGPT through supervised fine-tuning on PaveInstruct. The training objective minimizes the negative log-likelihood of target responses:
\begin{equation}
\label{cross_entropy_1}
\mathcal{L}(\theta) = -\sum_{i=1}^{N} \log p(r_{i,1}, ..., r_{i,T_i} | \mathcal{I}_i, \mathcal{Q}_i; \theta)
\end{equation}
where N is the total number of training samples, $r_{i,t}$ denotes the $t$-th token in the target response for sample $i$, and $T_i$ is the response length. The autoregressive factorization is:
\begin{equation}
\label{cross_entropy_2}
p(\mathcal{R}_i | \mathcal{I}_i, \mathcal{Q}_i; \theta) = \prod_{t=1}^{T_i} p(r_{i,t} | \mathcal{I}_i, \mathcal{Q}_i, r_{i,<t}; \theta)
\end{equation}

The language model's pre-trained knowledge of numerical reasoning and spatial relationships transfers to pavement-specific tasks. Our PaveInstruct dataset provides approximately 278,000 instruction-response pairs spanning 32 task types across five major categories, enabling PaveGPT to develop robust visual understanding that generalizes across task boundaries. The resulting foundation model seamlessly transitions between distress identification, condition estimation, and maintenance recommendations through natural language interaction.

\subsubsection{Overall Architecture}
\begin{figure}[H]
\centering
\includegraphics[scale=1.0]{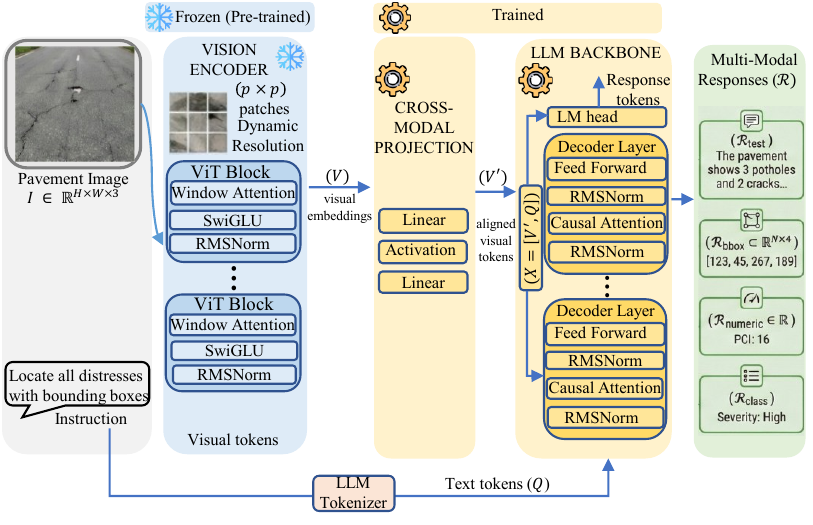}
\caption{PaveGPT architecture}
\label{fig:pavegpt_architecture}
\end{figure}

PaveGPT adopts the Qwen2.5-VL architecture, which follows a vision-language encoder-decoder design optimized for multimodal understanding. The model inherits three core components from Qwen2.5-VL: the vision transformer encoder, the large language model backbone, and the multimodal projection module. We adapt these components and conduct supervised fine-tuning on PaveInstruct to specialize the model for pavement infrastructure assessment. Figure~\ref{fig:pavegpt_architecture} shows the overall architecture for PaveGPT. 



\subsubsection{Vision Encoder}
The vision encoder $\mathcal{E}_v$ inherits the Vision Transformer (ViT) architecture from Qwen2.5-VL, which incorporates several optimizations for efficient processing. The encoder employs 2D-RoPE and window attention mechanisms that partition the image into non-overlapping windows for localized self-attention computation, reducing complexity while preserving spatial relationships crucial for distress localization.

Given an RGB pavement image $\mathcal{I} \in \mathbb{R}^{H \times W \times 3}$, the encoder applies dynamic resolution processing to preserve fine-grained details. The image is partitioned into patches of size $p \times p$, and each patch is linearly embedded to produce visual tokens:
\begin{equation}
\mathbf{V} = \mathcal{E}_v(\mathcal{I}) \in \mathbb{R}^{L_v \times d}
\label{eq:vision_encoder}
\end{equation}
where $L_v = \lfloor H/p \rfloor \times \lfloor W/p \rfloor$ is the number of patches and $d$ is the encoder's hidden dimension. The encoder uses SwiGLU activations and RMSNorm for layer normalization, following modern transformer design principles.

We initialize the vision encoder with Qwen2.5-VL's pre-trained weights and keep it frozen during fine-tuning. This preserves the strong general visual representations learned during Qwen2.5-VL's pre-training, which include low-level visual features (edges, textures, colors) and high-level concepts (objects, spatial relationships) that transfer effectively to pavement distress recognition. The frozen encoder provides stable visual features or cues that the trainable projection module learns to align with pavement-specific language.

\subsubsection{Cross-Modal Projection}
The multimodal projection module $\mathcal{P}$ bridges the vision encoder's output space and the language model's input space. This alignment is critical for enabling the language backbone to process visual information effectively alongside text instructions.

The projection is implemented as a learnable multilayer perceptron (MLP) that transforms visual embeddings $\mathbf{V}$ into the language model's embedding space:
\begin{equation}
\mathbf{V}' = \mathcal{P}(\mathbf{V}) = \text{MLP}(\mathbf{V}) \in \mathbb{R}^{L_v \times d_{llm}}
\label{eq:projection}
\end{equation}
where $d_{llm}$ is the language model's hidden dimension. The projected visual tokens $\mathbf{V}'$ are then concatenated with the tokenized instruction $\mathbf{Q}$ to form the input sequence:
\begin{equation}
\mathbf{X} = [\mathbf{V}', \mathbf{Q}] \in \mathbb{R}^{(L_v + L_q) \times d_{llm}}
\end{equation}

During training, we update the projection MLP parameters while keeping the vision encoder frozen, allowing the model to learn pavement-specific visual-language alignment without disrupting the rich pre-trained visual representations.


\subsubsection{Language Backbone}
The language backbone $\mathcal{M}_{llm}$ is a decoder-only transformer initialized from Qwen2.5-VL's large language model component. The backbone has been pre-trained on extensive general-domain corpora and multimodal instruction-following data, providing strong capabilities in natural language understanding, numerical reasoning, and structured output generation.

The language model processes the concatenated sequence $\mathbf{X} = [\mathbf{V}', \mathbf{Q}]$ through multiple transformer decoder layers with causal self-attention. At each generation step $t$, the model computes:
\begin{equation}
p(r_t | \mathcal{I}, \mathcal{Q}, r_{<t}; \theta) = \text{softmax}(\mathbf{W}_o \mathcal{M}_{llm}(\mathbf{X}, r_{<t}))
\end{equation}
where $\mathbf{W}_o$ projects the hidden states to vocabulary logits. This autoregressive generation enables the model to produce diverse output formats ranging from natural language descriptions to structured coordinates as token sequences.

Through supervised fine-tuning on PaveInstruct, the language backbone learns pavement engineering terminology, ASTM D6433 distress definitions, PCI calculation reasoning, and domain-specific spatial language for describing distress locations. The pre-trained numerical reasoning capabilities transfer effectively to tasks requiring coordinate generation and PCI estimation.

\subsubsection{Training and Optimization}
PaveGPT is fine-tuned on the PaveInstruct training split using the autoregressive objective defined in Equations~\ref{cross_entropy_1} and~\ref{cross_entropy_2}. During supervised fine-tuning, the vision tower remains frozen, while the multimodal projection MLP and the language backbone are updated. This setup keeps the general visual representations of Qwen2.5-VL intact and directs learning toward cross-modal alignment and pavement-specific reasoning.

We use the AdamW optimizer with separate learning rates for the two trainable modules: a small learning rate of $2 \times 10^{-7}$ for the language model parameters and a higher learning rate of $5 \times 10^{-4}$ for the multimodal MLP. The learning rate follows a cosine decay schedule with a warmup ratio of 0.03 and weight decay of 0.01. To stabilize optimization, we apply gradient clipping with a maximum norm of 1.0 and enable gradient checkpointing.

Fine-tuning runs for 10 epochs with a per-device batch size of 16 and gradient accumulation over 4 steps. The maximum sequence length is set to 8{,}192 tokens so that the model can handle long instructions and multi-turn examples. Training was conducted on the iTiger HPC cluster utilizing 8 NVIDIA H100 GPUs for 2 consecutive days, resulting in a cumulative total of 384 GPU hours \cite{sharif2025cultivatingmultidisciplinaryresearcheducation}. The main hyperparameters are summarized in Table~\ref{tab:pavegpt_hparams}.

\begin{table}[H]
    \centering
    \caption{Key hyperparameters used for fine-tuning PaveGPT.}
    \label{tab:pavegpt_hparams}
    \begin{tabular}{@{}ll@{}}
        \toprule
        \textbf{Parameter} & \textbf{Value} \\
        \midrule
        Training epochs & 10 \\
        Per-device train batch size & 16 \\
        Gradient accumulation steps & 4 \\
        Learning rate (LLM) & $2 \times 10^{-7}$ \\
        Learning rate (MM-MLP) & $5 \times 10^{-4}$ \\
        Weight decay & 0.01 \\
        Warmup ratio & 0.03 \\
        Learning-rate scheduler & Cosine decay \\
        Max sequence length & 8{,}192 tokens \\
        \bottomrule
    \end{tabular}
\end{table}

\subsection{Evaluation Metrics}

We evaluate PaveGPT across diverse pavement engineering tasks using a comprehensive framework of metrics tailored to different capability categories. Our evaluation strategy is organized into four primary groups: Spatial Grounding Metrics for object localization tasks, Structured Region Analysis Metrics for classification and parsing tasks, Reasoning and Comparative Assessment Metrics for complex analytical tasks, Vision-Language Generation Metrics for text generation tasks, and PCI Regression Metrics for pavement condition index assessment. Each metric group employs specialized evaluation approaches suited to the unique characteristics and requirements of its respective task category.

\subsubsection{Spatial Grounding Metrics}

Spatial grounding tasks assess the model's ability to localize pavement distresses by generating precise bounding box coordinates from natural language descriptions. These tasks include object detection grounding, referring expression comprehension, and spatial relationship analysis. We evaluate spatial grounding performance through metrics based on the overlap between predicted and ground truth bounding boxes.

\paragraph{\textbf{Intersection over Union}}
The Intersection over Union (IoU) \cite{everingham2010pascal} metric serves as the foundation for all spatial grounding evaluation. It quantifies the spatial overlap between a predicted bounding box and its corresponding ground truth annotation. For a predicted bounding box $\mathcal{B}_{\text{pred}} = [x_1^p, y_1^p, x_2^p, y_2^p]$ and ground truth box $\mathcal{B}_{\text{gt}} = [x_1^g, y_1^g, x_2^g, y_2^g]$, where $(x_1, y_1)$ denotes the top-left corner and $(x_2, y_2)$ denotes the bottom-right corner, we compute IoU as:

\begin{equation}
\text{IoU}(\mathcal{B}_{\text{pred}}, \mathcal{B}_{\text{gt}}) = \frac{\text{Area}(\mathcal{B}_{\text{pred}} \cap \mathcal{B}_{\text{gt}})}{\text{Area}(\mathcal{B}_{\text{pred}} \cup \mathcal{B}_{\text{gt}})}
\end{equation}

The intersection area is calculated as:

\begin{equation}
\text{Area}(\mathcal{B}_{\text{pred}} \cap \mathcal{B}_{\text{gt}}) = \max(0, x_{\min} - x_{\max}) \times \max(0, y_{\min} - y_{\max})
\end{equation}

where $x_{\min} = \min(x_2^p, x_2^g)$, $x_{\max} = \max(x_1^p, x_1^g)$, $y_{\min} = \min(y_2^p, y_2^g)$, and $y_{\max} = \max(y_1^p, y_1^g)$. The union area equals the sum of individual box areas minus the intersection area. IoU values range from 0 (no overlap) to 1 (perfect overlap), with values above 0.5 indicating successful localization and values above 0.7 representing precise spatial grounding suitable for professional applications.

\paragraph{\textbf{Detection Performance Metrics}}
Building on the IoU foundation, we compute standard detection metrics to assess localization quality comprehensively. Since our model generates bounding boxes as text without confidence scores, we use a fixed IoU threshold $\tau = 0.5$ to classify predictions. For each predicted box $\mathcal{B}_{\text{pred}}^i$ (where $i$ indexes predictions) and ground truth box $\mathcal{B}_{\text{gt}}^j$ (where $j$ indexes ground truth annotations), a predicted box is classified as a True Positive (TP) if there exists an unmatched ground truth box such that $\text{IoU}(\mathcal{B}_{\text{pred}}^i, \mathcal{B}_{\text{gt}}^j) \geq \tau$, with each ground truth box matched at most once. Predicted boxes that cannot be matched to any ground truth box with sufficient overlap are classified as False Positives (FP), representing spurious detections. Ground truth boxes that remain unmatched by any prediction are classified as False Negatives (FN), representing missed detections.

Using these classifications, we compute precision \cite{powers2011evaluation} as the proportion of correct localizations among all predictions:

\begin{equation}
\text{Precision} = \frac{\text{TP}}{\text{TP} + \text{FP}}
\end{equation}

Recall \cite{powers2011evaluation} measures the proportion of actual distresses successfully localized:

\begin{equation}
\text{Recall} = \frac{\text{TP}}{\text{TP} + \text{FN}}
\end{equation}

The F1-Score \cite{powers2011evaluation} harmonically combines precision and recall to provide a balanced performance measure:

\begin{equation}
\text{F1-Score} = 2 \times \frac{\text{Precision} \times \text{Recall}}{\text{Precision} + \text{Recall}}
\end{equation}

High precision indicates the model makes few spurious predictions, which is critical for avoiding unnecessary maintenance interventions. High recall ensures comprehensive distress detection with minimal omissions, which is essential for complete condition assessment. F1-scores above 0.7 indicate strong spatial grounding capabilities suitable for professional pavement inspection.

Finally, we compute the mean IoU across all matched prediction-ground truth pairs to assess average localization precision. Let $N_{\text{match}}$ denote the total number of successfully matched pairs. The mean IoU is then:

\begin{equation}
\text{Mean IoU} = \frac{1}{N_{\text{match}}}\sum_{i=1}^{N_{\text{match}}} \text{IoU}(\mathcal{B}_{\text{pred}}^i, \mathcal{B}_{\text{gt}}^i)
\end{equation}

This metric provides insight into localization quality beyond the binary detection metrics.

\subsubsection{Structured Region Analysis Metrics}

Region analysis tasks require the model to assess pavement conditions in a structured format, including distress type classification, severity level determination, and repair recommendation generation. The model generates free-text responses that must be parsed into structured fields before evaluation. We evaluate these tasks through field-specific accuracy metrics and format compliance measures.

\paragraph{\textbf{Field-Specific Classification Accuracy}}
For each structured field $f \in \\ \{\text{Distress, Severity, Repair}\}$, we compute classification accuracy by comparing predicted labels against ground truth annotations. Let $N$ denote the total number of evaluation samples, $\hat{y}_{i,f}$ denote the predicted label for field $f$ in sample $i$, and $y_{i,f}$ denote the corresponding ground truth label. The accuracy for field $f$ is:

\begin{equation}
\text{Accuracy}_f = \frac{1}{N} \sum_{i=1}^{N} \mathbb{I}(\text{match}_f(\hat{y}_{i,f}, y_{i,f}))
\end{equation}

where $\mathbb{I}(\cdot)$ is the indicator function returning 1 for correct predictions and 0 otherwise. The matching function $\text{match}_f(\cdot, \cdot)$ is field-dependent and defined as:

\begin{equation}
\text{match}_f(\hat{y}, y) = \begin{cases}
\mathbb{I}(\text{normalize}(\hat{y}) = \text{normalize}(y)) & \text{if } f = \text{Severity} \\
\mathbb{I}(\text{similarity}(\hat{y}, y) > 0.7) & \text{otherwise}
\end{cases}
\end{equation}

where $\text{normalize}(\cdot)$ converts text to lowercase and removes whitespace, and $\text{similarity}(\cdot, \cdot)$ computes fuzzy string similarity as detailed below. This field-specific matching strategy balances evaluation rigor with appropriate flexibility for each field type.

\paragraph{\textbf{Matching Strategies by Field Type}}
Different fields require different matching strategies to account for their distinct characteristics. For distress type and repair recommendation fields, we employ fuzzy string matching to accommodate minor textual variations while preserving semantic equivalence. The similarity between predicted and ground truth text is measured using the Levenshtein distance \cite{levenshtein1966binary}, which quantifies the minimum number of single-character edits (insertions, deletions, or substitutions) required to transform one string into another.

Formally, for strings $\hat{y}$ and $y$ of lengths $m = |\hat{y}|$ and $n = |y|$ respectively, the Levenshtein distance $\text{Lev}(i, j)$ is defined recursively as:

\begin{equation}
\text{Lev}(i, j) = \begin{cases}
\max(i, j) & \text{if } \min(i,j) = 0 \\
\min \begin{cases}
\text{Lev}(i-1, j) + 1 \\
\text{Lev}(i, j-1) + 1 \\
\text{Lev}(i-1, j-1) + \mathbb{I}(\hat{y}[i] \neq y[j])
\end{cases} & \text{otherwise}
\end{cases}
\end{equation}

where $i$ and $j$ index character positions in $\hat{y}$ and $y$ respectively, $\hat{y}[i]$ denotes the $i$-th character of $\hat{y}$, $y[j]$ denotes the $j$-th character of $y$, and $\mathbb{I}(\hat{y}[i] \neq y[j])$ equals 0 if characters match and 1 otherwise. The three operations in the minimum correspond to deletion (removing a character from $\hat{y}$), insertion (adding a character to $\hat{y}$), and substitution (replacing a character in $\hat{y}$). The final Levenshtein distance is $\text{Lev}(m, n)$.

We normalize this distance to obtain a similarity score bounded between 0 and 1:

\begin{equation}
\text{similarity}(\hat{y}, y) = 1 - \frac{\text{Lev}(m, n)}{\max(m, n)}
\end{equation}

A prediction is considered correct if $\text{similarity}(\hat{y}, y) > 0.7$. This threshold allows for minor phrasing differences (such as "longitudinal crack" versus "longitudinal cracking") while maintaining semantic accuracy.

In contrast, severity levels require exact matching after normalization to ensure consistent classification. After converting both prediction and ground truth to lowercase and removing whitespace, we apply strict equality:

\begin{equation}
\text{match}(\hat{y}_{i,\text{severity}}, y_{i,\text{severity}}) = \mathbb{I}(\text{normalize}(\hat{y}_{i,\text{severity}}) = \text{normalize}(y_{i,\text{severity}}))
\end{equation}

This strict approach is necessary because severity levels (Low, Medium, High) directly impact maintenance priority decisions and must be precisely classified.

\paragraph{Format Compliance Assessment}
Beyond content accuracy, the model must generate responses in the required structured format for automatic evaluation and system integration. The parsing success rate measures format compliance:
\begin{equation}
\text{Parsing Rate} = \frac{\sum_{i=1}^{N} \mathbb{I}(\text{parsable}(\hat{y}_i))}{N}
\end{equation}
where $N$ is the total number of evaluation samples, $\hat{y}_i$ is the generated free-text response for sample $i$, and $\text{parsable}(\hat{y}_i)$ returns true if all required fields were successfully extracted from the response. High parsing rates indicate the model reliably follows output specifications, which is essential for integration with pavement management systems.

\subsubsection{Reasoning and Comparative Assessment Metrics}

Complex reasoning tasks produce open-ended responses that require evaluation beyond simple string matching. These tasks include comparative grounding (comparing multiple distresses), chain-of-thought analysis (explaining reasoning steps), and maintenance decision-making (justifying treatment recommendations). We employ LLM-as-a-judge \cite{zheng2023judging} evaluation to assess the semantic correctness, reasoning quality, and technical soundness of these responses.

\paragraph{\textbf{LLM-as-a-Judge Evaluation Framework}}
We use a powerful language model (GPT-4o or Gemini 1.5 Pro) as an evaluator to assess response quality. For each evaluation sample $j$ (where $j \in \{1, \ldots, M\}$ and $M$ is the total number of samples evaluated through LLM judging), the judge evaluates the model's prediction $P_j$ against ground truth $G_j$ given question context $Q_j$, assigning a scalar score $S_j \in [1, 10]$ based on a detailed evaluation rubric. The rubric assesses five key dimensions: (1) factual correctness of technical claims, (2) logical coherence of reasoning chains, (3) completeness of analysis, (4) appropriate use of pavement engineering terminology, and (5) grounding in observable evidence. The mean judge score aggregates performance across all evaluated samples:

\begin{equation}
\text{Mean Judge Score} = \frac{1}{M}\sum_{j=1}^{M} S_j
\end{equation}

Scores of 6-8 indicate good technical reasoning quality, while scores above 8 represent excellent professional-grade analysis suitable for engineering decision support applications.

\paragraph{\textbf{Binary Success Assessment}}
To complement the continuous judge scores with a clear success threshold, we define a binary pass metric. A response "passes" if it achieves a score of 7 or higher, corresponding to "Good" quality or better according to our evaluation rubric. The pass rate measures the proportion of predictions meeting this professional quality standard:

\begin{equation}
\text{Pass Rate} = \frac{1}{M}\sum_{j=1}^{M} \mathbb{I}(S_j \geq 7)
\end{equation}

where $S_j$ is the judge score for sample $j$ and $\mathbb{I}(\cdot)$ is the indicator function. Pass rates above 70\% indicate the model consistently produces reasoning suitable for real-world pavement management, where technical accuracy and logical soundness are critical for safety and resource allocation decisions.

\paragraph{\textbf{Dimensional Quality Analysis}}
Beyond aggregate scores, we evaluate each of the five quality dimensions separately to identify specific strengths and weaknesses. These dimensions assess: (1) Factual Accuracy, (2) Logical Coherence, (3) Technical Terminology, (4) Evidence Grounding, and (5) Completeness. For each dimension $d$ and each sample $j$, the judge assigns a dimension-specific score $S_{j,d} \in [1, 10]$. The mean score for dimension $d$ is:

\begin{equation}
\text{Mean Score}_d = \frac{1}{M}\sum_{j=1}^{M} S_{j,d}
\end{equation}

This dimensional breakdown provides actionable insights for model improvement by revealing which aspects of reasoning require refinement.

\subsubsection{Vision-Language Generation Metrics}

For tasks involving natural language generation,specifically pavement captioning and visual question answering, we employ established vision-language metrics. These metrics assess the quality of generated text through n-gram matching and semantic similarity with reference text.

\paragraph{Captioning Quality Metrics.}
We evaluate caption quality using three complementary metrics that capture different aspects of text similarity.

\textbf{BLEU-4.} BLEU (Bilingual Evaluation Understudy) \cite{papineni2002bleu} measures n-gram precision between generated and reference captions. For n-gram order $n \in \{1, 2, 3, 4\}$, the precision $P_n$ is computed as:
\begin{equation}
P_n = \frac{\sum_{\text{ngram} \in C} \min(\text{Count}_{\text{clip}}(\text{ngram}), \text{Count}_{\text{ref}}(\text{ngram}))}{\sum_{\text{ngram} \in C} \text{Count}(\text{ngram})}
\end{equation}
where $C$ denotes the candidate caption, $\text{Count}(\text{ngram})$ is the count of n-gram occurrences in the candidate, $\text{Count}_{\text{ref}}(\text{ngram})$ is the maximum count of the n-gram in any reference caption, and $\text{Count}_{\text{clip}}$ clips n-gram counts to their maximum reference occurrence. To prevent artificially high scores from overly short outputs, we apply a brevity penalty:
\begin{equation}
\text{BP} = \begin{cases}
1 & \text{if } c > r \\
e^{(1 - r/c)} & \text{if } c \leq r
\end{cases}
\end{equation}
where $c$ is the candidate caption length (in words) and $r$ is the reference caption length. The BLEU-4 score combines precision across n-gram orders 1 through 4:
\begin{equation}
\text{BLEU-4} = \text{BP} \cdot \exp\left(\frac{1}{4}\sum_{n=1}^{4} \log P_n\right)
\end{equation}
BLEU-4 scores above 0.3 are considered good for technical domain text generation.

\textbf{ROUGE-L.} ROUGE-L \cite{lin2004rouge} measures the longest common subsequence (LCS) between candidate and reference text, capturing sentence-level structure preservation. Let $X$ denote the candidate caption and $Y$ denote a reference caption, with lengths $|X|$ and $|Y|$ measured in words. Let $\text{LCS}(X, Y)$ denote the length of the longest common subsequence between $X$ and $Y$. We compute LCS-based recall and precision:
\begin{equation}
R_{\text{lcs}} = \frac{\text{LCS}(X, Y)}{|Y|}, \quad P_{\text{lcs}} = \frac{\text{LCS}(X, Y)}{|X|}
\end{equation}
These are combined into an F-measure with recall emphasis (using parameter $\beta = 1.2$):
\begin{equation}
\text{ROUGE-L} = \frac{(1 + \beta^2) R_{\text{lcs}} P_{\text{lcs}}}{R_{\text{lcs}} + \beta^2 P_{\text{lcs}}}
\end{equation}
Scores above 0.4 indicate strong structural alignment with reference descriptions.

\textbf{CIDEr.} CIDEr (Consensus-based Image Description Evaluation) \cite{vedantam2015cider} emphasizes distinctive n-grams through TF-IDF weighting, rewarding descriptions that capture salient details. For n-gram $\omega_k$ (where $k$ indexes all possible n-grams) in sentence $s_{ij}$ (where $i$ indexes images and $j$ indexes reference sentences for image $i$), the TF-IDF weight $g_k(s_{ij})$ is:
\begin{equation}
g_k(s_{ij}) = \frac{h_k(s_{ij})}{\sum_{\omega_l} h_l(s_{ij})} \log\left(\frac{|I|}{\sum_{I_p \in I} \min(1, \sum_q h_k(s_{pq}))}\right)
\end{equation}
where $h_k(s_{ij})$ counts occurrences of n-gram $\omega_k$ in sentence $s_{ij}$, $I$ is the set of all images in the dataset with $|I|$ denoting the total number of images, $I_p$ indexes each image in the dataset, $q$ indexes reference sentences for image $p$, and the summation $\sum_{\omega_l}$ runs over all n-grams in the sentence. The CIDEr score for candidate caption $c_i$ (for image $i$) and reference set $S_i = \{s_{i1}, s_{i2}, \ldots, s_{im}\}$ (containing $m$ reference captions) averages cosine similarities of TF-IDF vectors:
\begin{equation}
\text{CIDEr}_n(c_i, S_i) = \frac{1}{m}\sum_{j=1}^{m} \frac{\mathbf{g}^n(c_i) \cdot \mathbf{g}^n(s_{ij})}{\|\mathbf{g}^n(c_i)\| \|\mathbf{g}^n(s_{ij})\|}
\end{equation}
where $\mathbf{g}^n(c_i)$ is the TF-IDF vector for n-grams of length $n$ in candidate caption $c_i$, $\mathbf{g}^n(s_{ij})$ is the TF-IDF vector for reference caption $s_{ij}$, and $\|\cdot\|$ denotes the vector norm. The final score averages over multiple n-gram orders (typically $n \in \{1, 2, 3, 4\}$), with higher values indicating better consensus with human descriptions.

\paragraph{\textbf{Question Answering Accuracy Metrics}}
For visual question answering tasks, we employ two accuracy metrics with different stringency levels. Let $N$ denote the total number of VQA evaluation samples, $\hat{y}_i$ denote the predicted answer for question $i$, and $y_i$ denote the ground truth answer. Exact match accuracy measures perfect agreement after text normalization:

\begin{equation}
\text{Exact Match} = \frac{1}{N}\sum_{i=1}^{N} \mathbb{I}[\text{normalize}(\hat{y}_i) = \text{normalize}(y_i)]
\end{equation}

where normalization includes converting to lowercase, removing articles (a, an, the), and stripping punctuation. This strict metric ensures precise answer correctness.

To account for semantically equivalent but textually different answers, we also compute relaxed accuracy. For each sample $i$, we define a soft matching indicator:
\begin{equation}
\text{SoftMatch}_i = \mathbb{I}[\text{similarity}(\hat{y}_i, y_i) > 0.8] \vee \mathbb{I}[\hat{y}_i \subseteq y_i \vee y_i \subseteq \hat{y}_i]
\end{equation}
where $\text{similarity}(\cdot, \cdot)$ is the normalized Levenshtein distance defined in Equation~10, $\subseteq$ denotes substring containment, and $\vee$ denotes logical OR. The relaxed accuracy aggregates these matches:
\begin{equation}
\text{Relaxed Accuracy} = \frac{1}{N}\sum_{i=1}^{N} \text{SoftMatch}_i
\end{equation}

This metric better captures semantic correctness in technical question answering, where multiple valid formulations exist for the same answer (such as "yes, severe alligator cracking" and "yes").

\subsubsection{PCI Regression Metrics}
PCI estimation is evaluated using standard regression metrics that quantify prediction error magnitude and distribution.

\paragraph{\textbf{Mean Absolute Error (MAE)}}
MAE measures the average absolute deviation between predicted and actual PCI values:

\begin{equation}
\text{MAE} = \frac{1}{N}\sum_{i=1}^{N} |\hat{y}_i - y_i|
\end{equation}

where $\hat{y}_i$ is the predicted PCI value and $y_i$ is the ground truth PCI value for sample $i$. MAE provides an interpretable measure of average error magnitude in the same units as PCI (0-100 scale), with lower values indicating better performance. For PCI estimation, MAE below 10 points is considered good accuracy.

\paragraph{\textbf{Mean Squared Error (MSE)}}
MSE quantifies the average squared deviation, penalizing larger errors more heavily:

\begin{equation}
\text{MSE} = \frac{1}{N}\sum_{i=1}^{N} (\hat{y}_i - y_i)^2
\end{equation}

The squared term makes MSE more sensitive to outliers than MAE, providing insight into prediction variance. The Root Mean Squared Error (RMSE) is often reported for interpretability:

\begin{equation}
\text{RMSE} = \sqrt{\text{MSE}}
\end{equation}

Lower MSE and RMSE values indicate better model performance, with RMSE below 15 considered acceptable for practical PCI prediction applications.

\section{Results and Discussion}
This section presents comprehensive evaluation results demonstrating the effectiveness of supervised instruction fine-tuning for pavement assessment. Quantitative benchmarks compare PaveGPT and fine-tuned baseline models across perception, understanding, and reasoning tasks in both zero-shot and trained settings. Qualitative analysis examines model outputs across diverse task types, while computational efficiency metrics assess practical deployment feasibility.

\subsection{Quantitative Analysis}

\begin{figure}[H]
\centering
\includegraphics[width=\columnwidth]{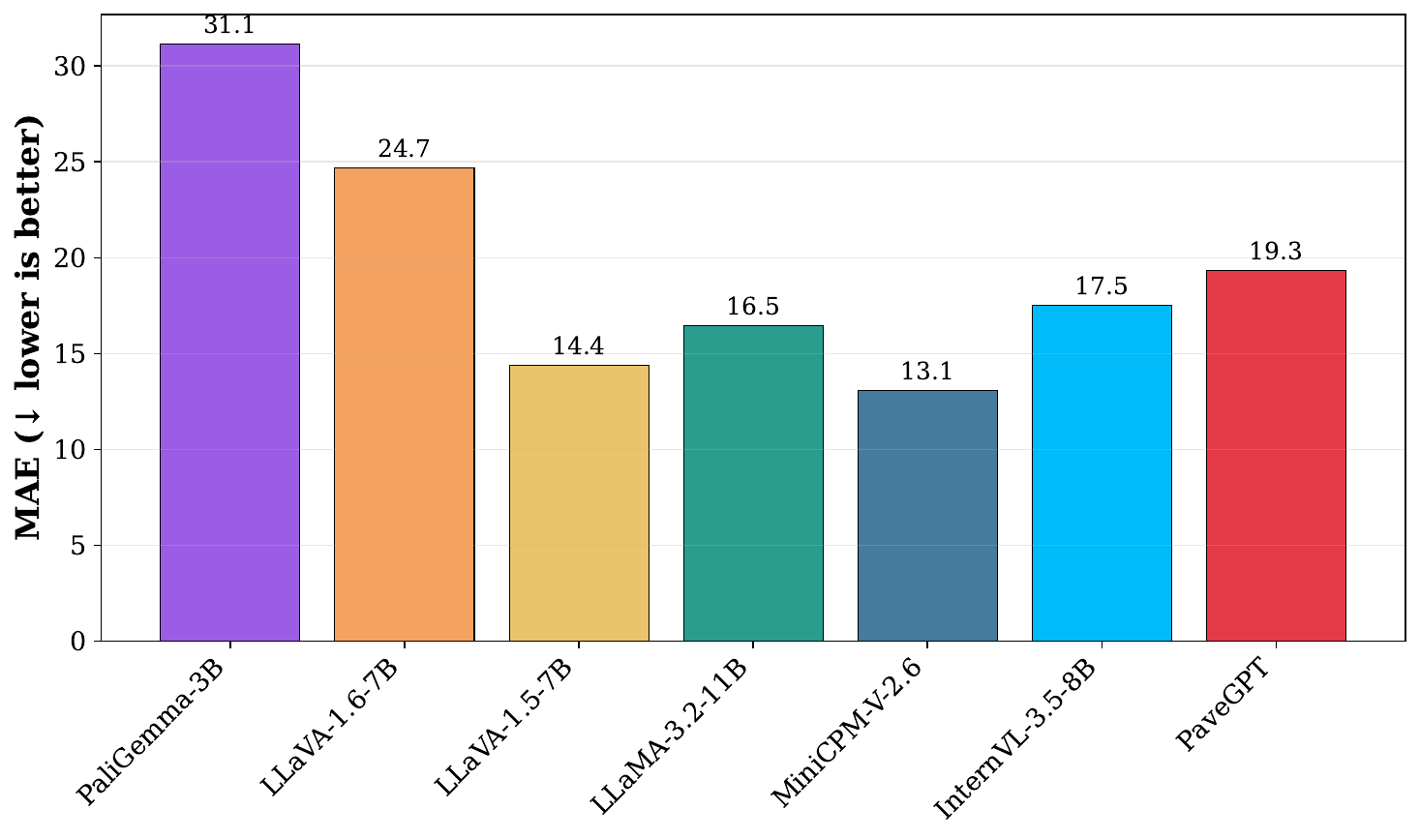}
\caption{PCI prediction Mean Absolute Error (MAE) comparison. Lower values indicate better performance.}
\label{fig:pci_mae}
\end{figure}

Table~\ref{tab:unified_results} reports results across a comprehensive suite of perception, understanding, and explanatory tasks, allowing a unified evaluation of different state-of-the-art VLMs within the context of pavement condition assessment. In the zero-shot setting, most models fail to produce meaningful outputs for tasks involving structured predictions such as distress localization, severity classification, and PCI estimation. In many cases, outputs are entirely missing or do not follow expected formats, making them impossible to evaluate ( as indicated by the prevalence of ``--'' in zero-shot rows). Even where numerical values are generated, they tend to be unreliable and inconsistent with domain expectations. For example, PCI predictions from LLaMA-3.2 and MiniCPM in zero-shot mode yielded extremely high error values, highlighting the limitations of using existing general purpose VLMs alone for specialized technical domains. These failures reinforce the motivation for this work, which emphasizes the need for targeted supervision strategies that guide models to understand and follow the conventions, formats, and reasoning styles used in professional pavement assessment. Figure~\ref{fig:pci_mae} compares the PCI prediction accuracy across models, where MiniCPM-V-2.6 achieves the lowest MAE of 13.1, followed by LLaVA-1.5-7B at 14.4.

\begin{figure}[H]
\centering
\includegraphics[width=1.0\columnwidth]{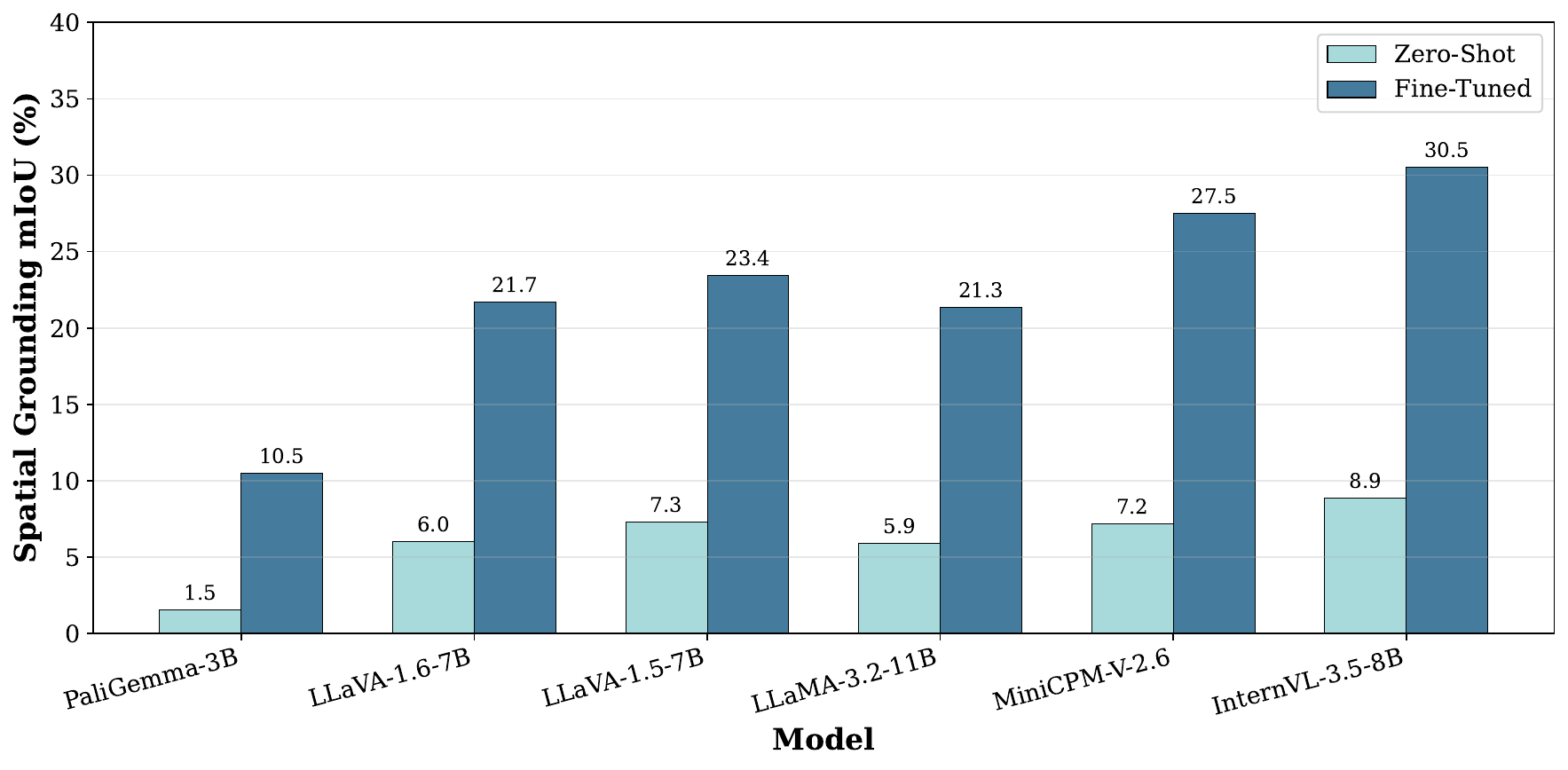}
\caption{Zero-shot vs. fine-tuned performance on spatial grounding (mIoU). Fine-tuning on PaveInstruct yields substantial improvements across all VLMs.}
\label{fig:zs_vs_ft}
\end{figure}

Supervised fine-tuning on the PaveInstruct dataset leads to consistent and substantial performance improvements across all models and task categories. The dataset's rich combination of instruction types, annotation formats, and task diversity allows models to learn how to respond accurately in a wide range of scenarios. On perception tasks, models like InternVL and MiniCPM show strong improvements in both regional distress understanding and spatial grounding. InternVL achieves the highest mIoU and box accuracy scores, indicating precise localization capabilities, while MiniCPM performs best on severity classification and region-level distress detection. Figure~\ref{fig:zs_vs_ft} shows the substantial improvement in spatial grounding performance (mIoU) when VLMs are fine-tuned on PaveInstruct compared to their zero-shot baselines.

\begin{figure}[H]
\centering
\includegraphics[width=1.0\columnwidth]{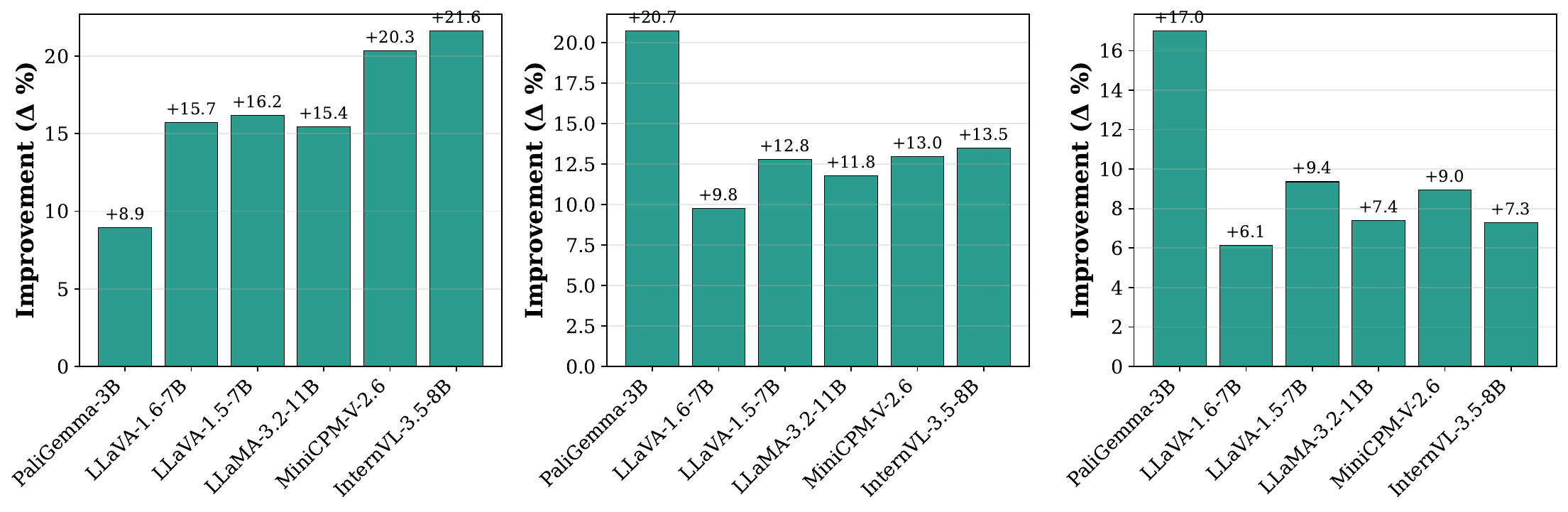}
\caption{Performance improvement ($\Delta$\%) from fine-tuning on PaveInstruct across spatial grounding, reasoning, and captioning tasks compared to zero-shot settings. }
\label{fig:performance_improvement}
\end{figure}

In understanding tasks, MiniCPM also excels at question answering and PCI estimation, achieving the lowest prediction error and the highest correlation with ground truth PCI values. These results reflect the effectiveness of instruction tuning not just in boosting raw accuracy, but also in helping models produce well-structured, interpretable, and technically valid outputs. Performance gains are not restricted to a single architecture; both smaller and larger models benefit significantly, indicating that the improvements are not simply due to model scale, but rather the relevance and quality of the task supervision provided. Figure~\ref{fig:performance_improvement} illustrates the performance gains achieved by fine-tuning on PaveInstruct across spatial grounding, reasoning, and captioning tasks, with all models showing consistent improvements.

\begin{figure}[H]
\centering
\includegraphics[width=1.0\columnwidth]{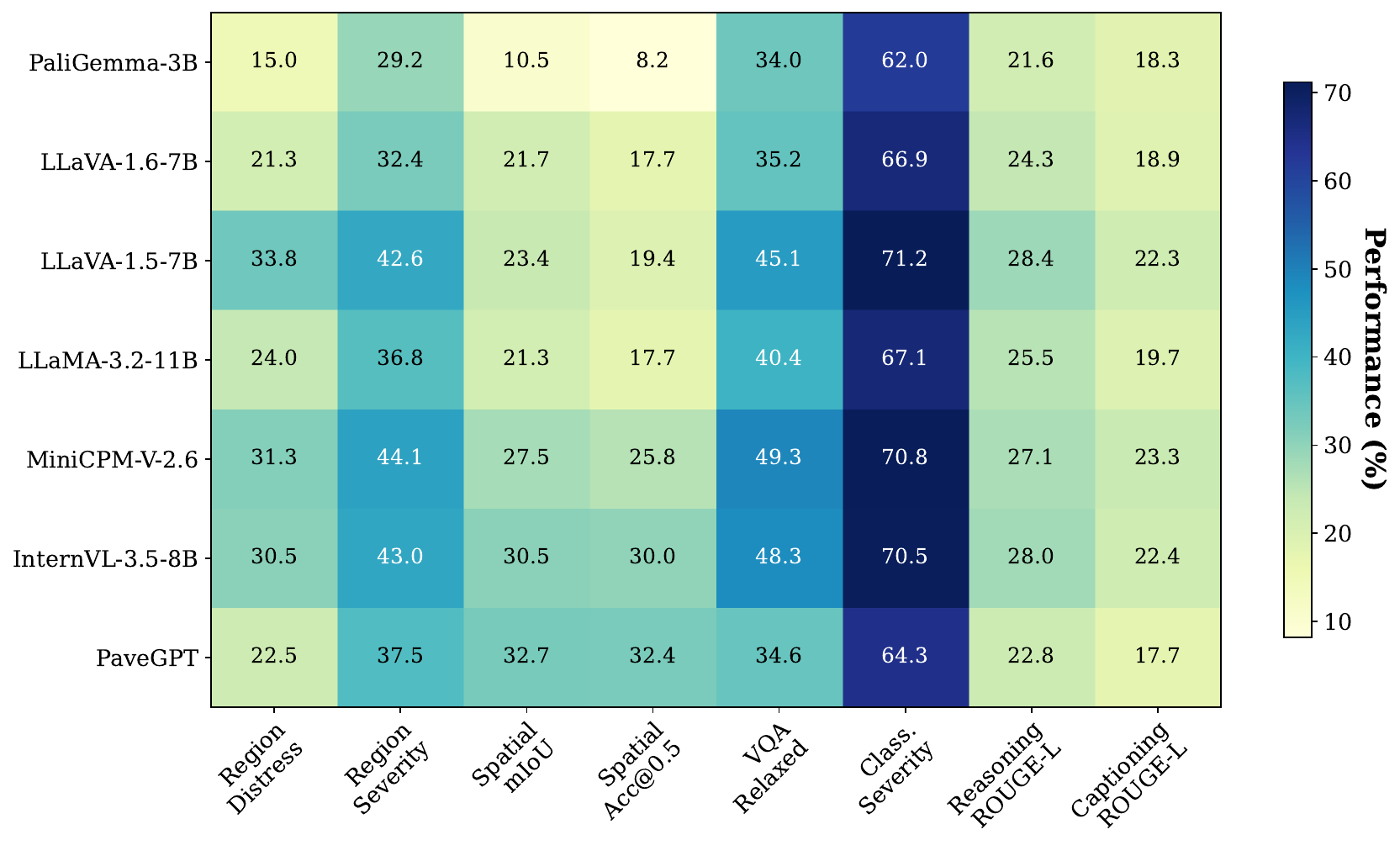}
\caption{Heatmap of fine-tuned model performance across evaluation metrics. Darker colors indicate higher scores.}
\label{fig:heatmap}
\end{figure}

In explanatory tasks, which include pavement captioning, models trained on PaveInstruct are able to generate more fluent, relevant, and context-sensitive outputs. MiniCPM achieves the highest scores across all generation metrics, including BLEU-4, ROUGE-L, and CIDEr, and also ranks highest in human-judged reasoning ability. These gains suggest that exposure to domain-specific language patterns helps models develop stronger conversational and explanatory abilities. While PaveGPT does not lead in overall scores, it achieves the best performance in spatial localization, reflecting its focus on integrating vision and structure-aware output generation. Taken together, these findings support the overall goal of the paper, which is to move beyond narrow, single-task models and instead enable general-purpose VLMs to handle the full spectrum of tasks encountered in real-world pavement evaluation workflows. Instruction tuning, when built around authentic domain needs and standardized outputs, is key to unlocking this capability. Figure~\ref{fig:heatmap} presents a comprehensive heatmap visualization of model performance across all evaluation metrics, enabling direct comparison of relative strengths and weaknesses (darker values indicates stronger scores).

\begin{table}[H]
\centering
\caption{Benchmark results comparing state-of-the-art VLMs on pavement condition assessment. Zero-shot (gray rows) shows baseline performance; Trained (white rows) shows performance after fine-tuning on PaveInstruct. Perception evaluates distress detection and spatial localization; Understanding covers VQA and PCI score prediction; Explanatory assesses caption quality and reasoning ability via LLM-based evaluation. Best in \textbf{bold}, second-best \underline{underlined}. $\uparrow$/$\downarrow$: higher/lower is better. \textbf{--}: not evaluable in zero-shot.}
\label{tab:unified_results}
\renewcommand{\arraystretch}{1.4}
\resizebox{\textwidth}{!}{%
\begin{tabular}{lc ccc cc c cc ccc ccc cc}
\toprule
& & \multicolumn{6}{c}{\textbf{Perception}} & \multicolumn{5}{c}{\textbf{Understanding}} & \multicolumn{5}{c}{\textbf{Explanatory}} \\
\cmidrule(lr){3-8} \cmidrule(lr){9-13} \cmidrule(lr){14-18}
& & \multicolumn{3}{c}{Region Analysis} & \multicolumn{2}{c}{Spatial Grounding} & Classification & \multicolumn{2}{c}{VQA} & \multicolumn{3}{c}{PCI Prediction} & \multicolumn{3}{c}{Captioning} & \multicolumn{2}{c}{Reasoning} \\
\cmidrule(lr){3-5} \cmidrule(lr){6-7} \cmidrule(lr){8-8} \cmidrule(lr){9-10} \cmidrule(lr){11-13} \cmidrule(lr){14-16} \cmidrule(lr){17-18}
\textbf{Method} & \textbf{Setting} & 
Distress$\uparrow$ & Severity$\uparrow$ & Repair$\uparrow$ & 
mIoU$\uparrow$ & Accuracy$\uparrow$ & 
Severity$\uparrow$ & 
\makecell{Exact\\Accuracy$\uparrow$} & \makecell{Relaxed\\Accuracy$\uparrow$} & 
MAE$\downarrow$ & RMSE$\downarrow$ & R$^2$$\uparrow$ & 
BLEU-4$\uparrow$ & ROUGE-L$\uparrow$ & CIDEr$\uparrow$ & 
\makecell{Judge\\Score$\uparrow$} & \makecell{Pass\\Rate$\uparrow$} \\
\midrule
\rowcolor{lightgray}
PaliGemma-3B & Zero-shot & -- & -- & -- & 1.55 & 1.48 & -- & -- & -- & -- & -- & -- & 0.00 & 1.30 & 0.00 & 6.38 & 62.5 \\
PaliGemma-3B & Trained & 15.00 & 29.16 & 1.71 & 10.47 & 8.17 & 62.04 & 11.81 & 33.96 & 31.13 & 42.84 & -0.73 & 6.04 & 18.30 & 8.05 & 5.12 & 30.0 \\
\addlinespace[0.3em]
\rowcolor{lightgray}
LLaVA-1.5-7B & Zero-shot & -- & -- & -- & 7.29 & 7.28 & -- & -- & -- & -- & -- & -- & 4.13 & 12.90 & 2.39 & 4.53 & 30.0 \\
LLaVA-1.5-7B & Trained & \textbf{33.76} & 42.55 & \underline{4.55} & 23.45 & 19.39 & \textbf{71.20} & 17.35 & 45.11 & \underline{14.37} & \underline{22.27} & \underline{0.54} & \textbf{10.08} & 22.25 & 15.50 & 6.56 & 56.0 \\
\addlinespace[0.3em]
\rowcolor{lightgray}
LLaVA-1.6-7B & Zero-shot & -- & -- & -- & 6.02 & 6.00 & -- & -- & -- & -- & -- & -- & 1.82 & 12.75 & 0.84 & 4.56 & 18.0 \\
LLaVA-1.6-7B & Trained & 21.25 & 32.37 & 1.14 & 21.73 & 17.72 & 66.87 & 12.38 & 35.18 & 24.68 & 37.22 & -0.31 & 5.83 & 18.89 & 4.10 & \underline{6.96} & 56.0 \\
\addlinespace[0.3em]
\rowcolor{lightgray}
LLaMA-3.2-11B & Zero-shot & -- & -- & -- & 5.89 & 5.81 & -- & -- & -- & 65.86 & 96.78 & -- & 2.40 & 12.26 & 1.06 & 3.66 & 14.0 \\
LLaMA-3.2-11B & Trained & 24.04 & 36.76 & 2.79 & 21.33 & 17.72 & 67.13 & 17.83 & 40.39 & 16.46 & 24.11 & 0.46 & 7.67 & 19.66 & 8.26 & \underline{6.96} & \underline{58.0} \\
\addlinespace[0.3em]
\rowcolor{lightgray}
MiniCPM-V-2.6 & Zero-shot & -- & -- & -- & 7.19 & 7.09 & -- & -- & -- & 20.85 & 31.20 & -- & 2.74 & 14.31 & 1.24 & 5.71 & 39.3 \\
MiniCPM-V-2.6 & Trained & \underline{31.28} & \textbf{44.05} & 3.88 & 27.52 & 25.79 & \underline{70.82} & \underline{21.58} & \textbf{49.27} & \textbf{13.07} & \textbf{21.47} & \textbf{0.57} & \textbf{10.08} & \textbf{23.26} & \textbf{19.60} & \textbf{7.28} & \textbf{72.0} \\
\addlinespace[0.3em]
\rowcolor{lightgray}
InternVL-3.5-8B & Zero-shot & -- & -- & -- & 8.89 & 8.86 & -- & -- & -- & 31.59 & 38.49 & -- & 2.73 & 15.15 & 1.95 & 4.86 & 20.0 \\
InternVL-3.5-8B & Trained & 30.46 & \underline{42.97} & \textbf{4.76} & \underline{30.52} & \underline{30.02} & 70.53 & \textbf{22.72} & \underline{48.29} & 17.50 & 26.35 & 0.35 & \underline{10.00} & \underline{22.44} & \underline{16.20} & 6.86 & \underline{58.0} \\
\addlinespace[0.3em]
\textbf{PaveGPT-7B} & Trained & 22.54 & 37.54 & 2.74 & \textbf{32.68} & \textbf{32.38} & 64.30 & 15.07 & 34.61 & 19.32 & 27.14 & 0.31 & 6.21 & 17.68 & 6.33 & 6.14 & 40.0 \\
\bottomrule
\end{tabular}%
}
\renewcommand{\arraystretch}{1.0}
\end{table}

\begin{figure}[H]
\centering
\includegraphics[width=\columnwidth]{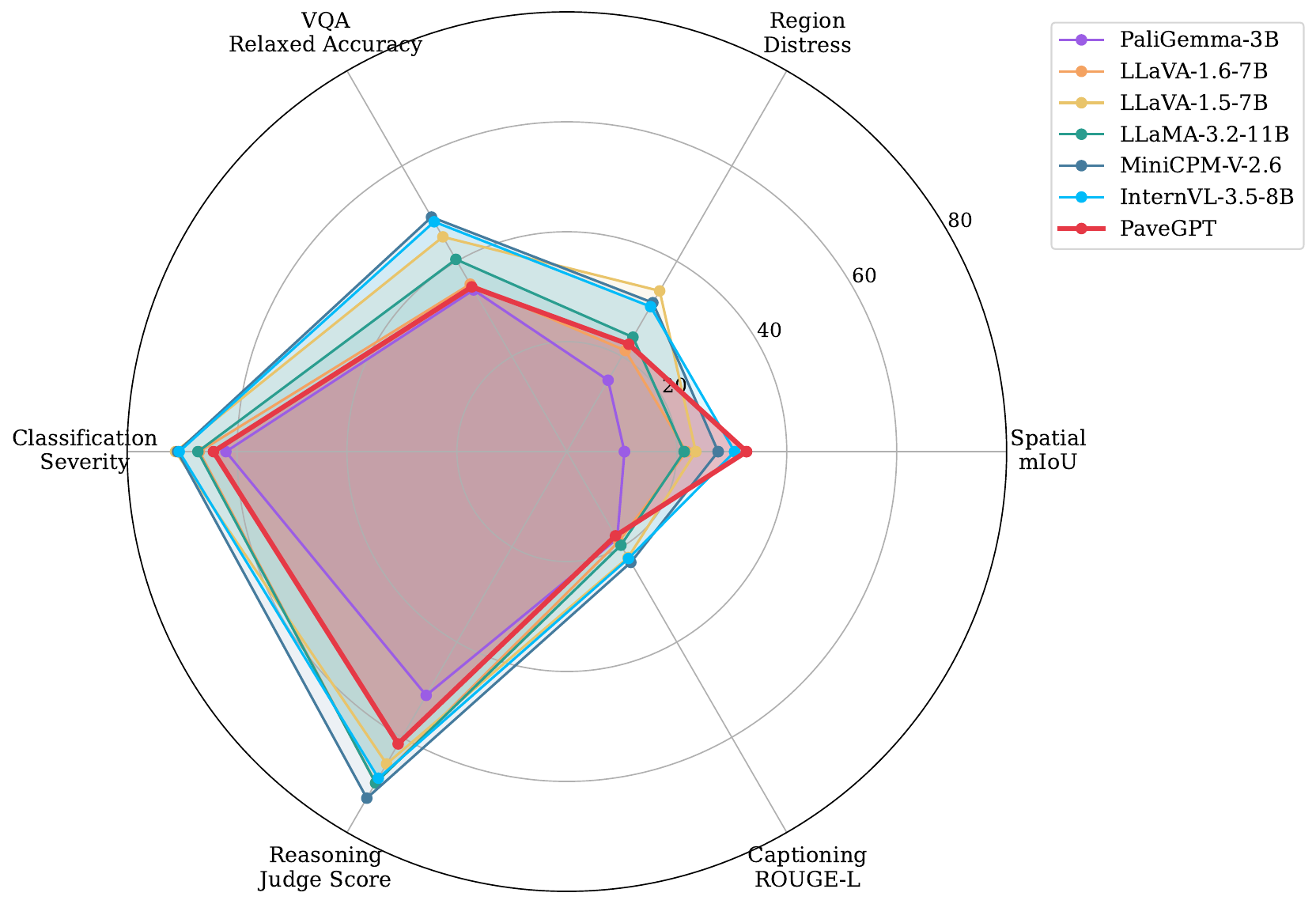}
\caption{Radar chart comparing fine-tuned VLMs across six key evaluation metrics on PaveInstruct.}
\label{fig:radar_chart}
\end{figure}

Figure~\ref{fig:radar_chart} provides a multi-dimensional comparison of fine-tuned VLMs across six key metrics, highlighting that different models exhibit distinct performance trade-offs.

\subsection{Qualitative Results}
\subsubsection{Zero-Shot vs. Fine-Tuned Performance}
\begin{figure}[H]
\centering
\includegraphics[scale=0.8]{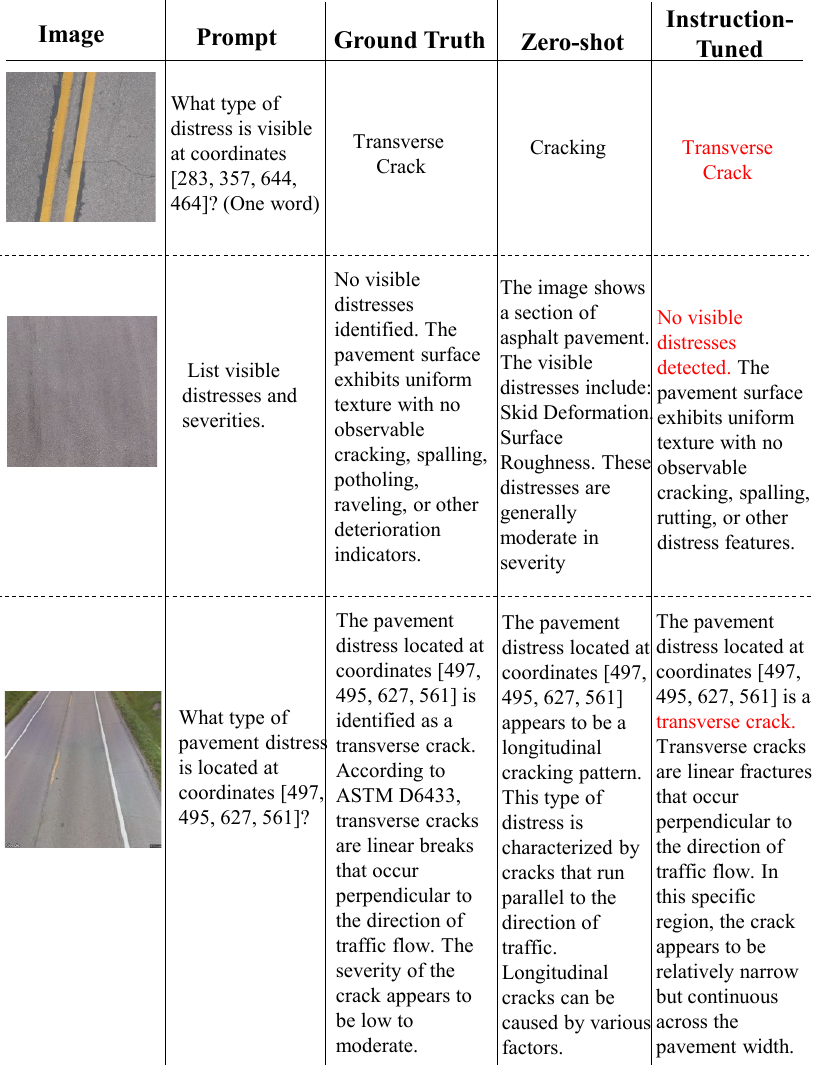}
\caption{Comparison of InternVL-3.5-8B performance before and after fine-tuning on PaveInstruct across diverse task types: one-word classification, distress enumeration, and coordinate-based identification. Red highlights indicate correct alignment with ground truth, demonstrating substantial improvements from domain-specific instruction tuning.}
\label{fig:zeroshot_vs_finetuned}
\end{figure}

To complement the quantitative analysis, we begin with a comparison between zero-shot and fine-tuned model behavior. Figure~\ref{fig:zeroshot_vs_finetuned} shows InternVL-3.5-8B's predictions before and after tuning on PaveInstruct, spanning three task types: distress classification, distress listing, and coordinate-based distress localization. In the zero-shot setting, the model produces vague or incorrect outputs. For instance, describing the pavement as having “uniform texture” despite the presence of visible transverse cracking. It also misidentifies distress types and fails to follow expected response formats. After fine-tuning, the model correctly detects the transverse crack and lists relevant distresses. These improvements illustrate how instruction tuning enables models to internalize pavement-specific vocabulary, reasoning logic, and structured response formats that are not present in general-domain pretraining.

\subsubsection{Fine-Tuned Model Comparisons Across Tasks}
Building on the improvements seen after fine-tuning, we now take a closer look at how PaveGPT and other VLMs perform across a range of pavement assessment tasks. All models here have been trained on PaveInstruct and are evaluated on four instruction-following tasks: distress listing, visual question answering (VQA), PCI estimation with justification, and spatial grounding. Each task highlights a different aspect of pavement reasoning, including semantic identification, diagnostic decision making, numeric assessment, and spatial localization. These examples illustrate how domain-aligned instruction tuning affects output structure, technical accuracy, and interpretability.

\begin{figure}[H]
\centering
\includegraphics[scale=0.8]{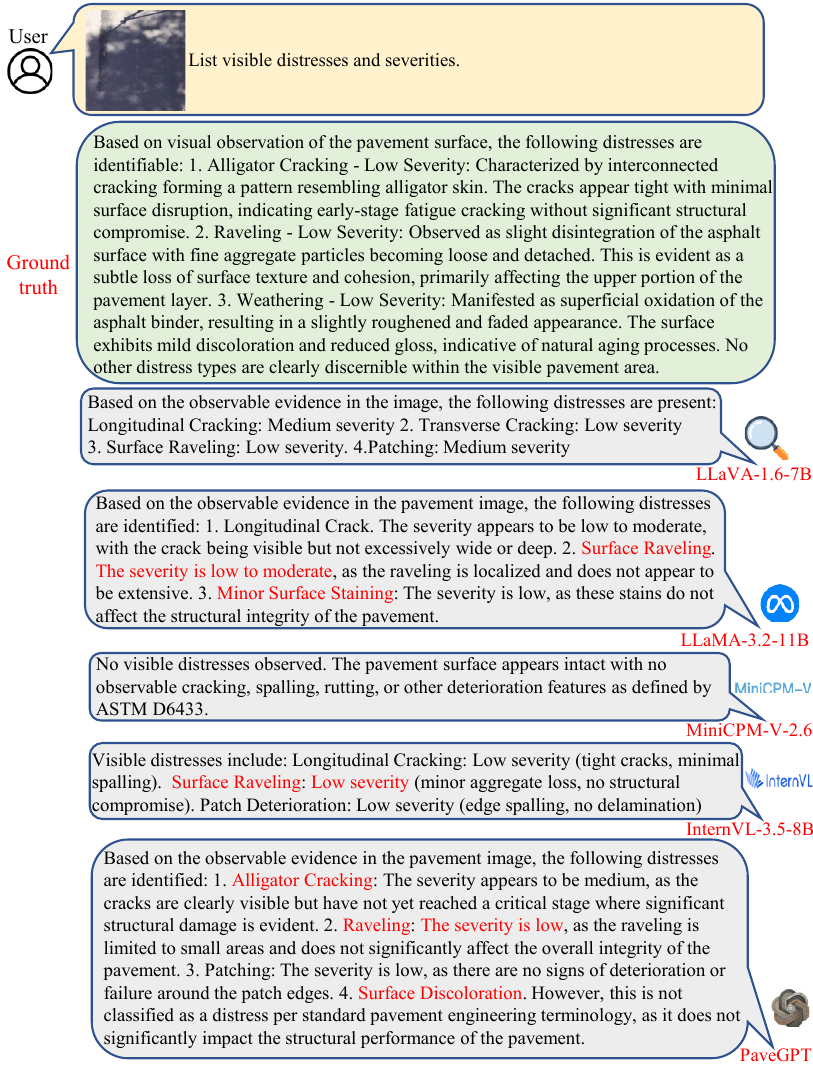}
\caption{Qualitative comparison of VLM responses for distress listing task. Red highlights indicate correct matches with ground truth.}
\label{fig:distress_listing}
\end{figure}

We first examine distress listing results, where models are asked to identify all visible distresses and assess their severity. As shown in Figure~\ref{fig:distress_listing}, PaveGPT and MiniCPM-V-2.6 generate detailed and well-structured outputs, correctly identifying multiple distress types such as alligator cracking, raveling, and patching. PaveGPT further distinguishes between structural and surface-level conditions, using terminology consistent with ASTM D6433. InternVL also detects major distresses, but introduces a non-standard term, surface discoloration, which reflects partial confusion between visual appearance and engineering-defined distress categories. LLaVA-1.6 produces accurate but less complete listings, while LLaMA-3.2 fails to identify any distress. These differences indicate that instruction tuning improves multi-class recognition and technical vocabulary usage, although sensitivity to subtle conditions remains model-dependent.

\begin{figure}[H]
\centering
\includegraphics[scale=0.8]{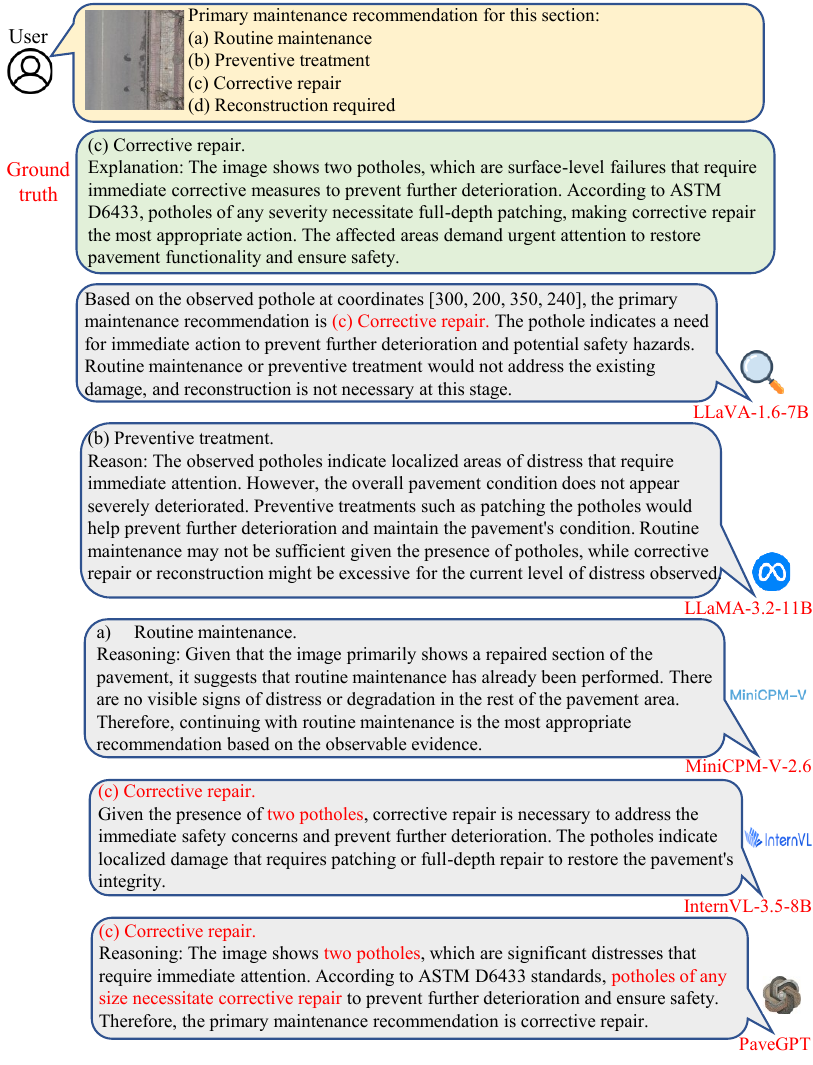}
\caption{Results on multiple-choice VQA for maintenance recommendation. Red highlights indicate correct matches with ground truth.}
\label{fig:vqa_maintenance}
\end{figure}

We next analyze maintenance recommendation through a multiple-choice VQA task. The ground truth action is corrective repair, based on the presence of potholes. Figure~\ref{fig:vqa_maintenance} shows that PaveGPT, MiniCPM, and InternVL correctly select this option and justify their choice by referencing safety concerns and ASTM D6433 guidance. Their explanations clearly link the observed distress type to the required intervention. In contrast, LLaVA-1.6 selects preventive treatment, suggesting difficulty in mapping potholes to the appropriate maintenance category. LLaMA-3.2 again underperforms, choosing routine maintenance with minimal reasoning. These examples show that instruction tuning improves not only answer accuracy, but also the logical connection between visual evidence and engineering decisions.

\begin{figure}[H]
\centering
\includegraphics[scale=0.8]{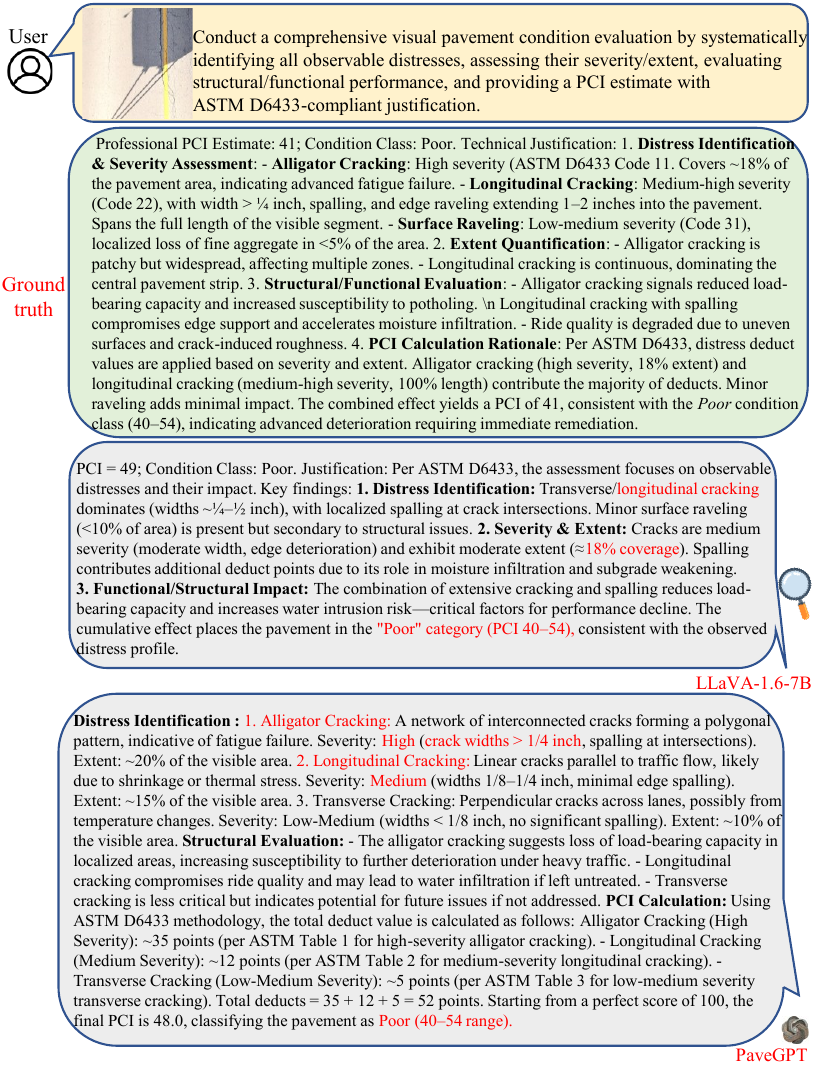}
\caption{Qualitative comparison on PCI estimation with ASTM D6433-compliant reasoning. Red highlights indicate correct matches.}
\label{fig:pci_estimation}
\end{figure}

PCI estimation provides a deeper view into structured numeric reasoning combined with visual assessment. As illustrated in Figure~\ref{fig:pci_estimation}, PaveGPT outputs a PCI value of 41 and provides a step-by-step justification that references distress types, severity levels, extent, and deduct values consistent with ASTM D6433. The explanation clearly separates major structural distresses from minor surface defects and explains their relative impact on the final score. LLaVA-1.6 also produces a compliant explanation and estimates a PCI of 49, which falls within the correct condition class range. Although the numeric values differ slightly, both outputs align with the ground truth condition rating of Poor. Other models either fail to generate a numeric PCI or do not provide a structured justification, indicating that PCI estimation remains challenging without explicit supervision on reasoning format and standards.

\begin{figure}[H]
\centering
\includegraphics[scale=0.8]{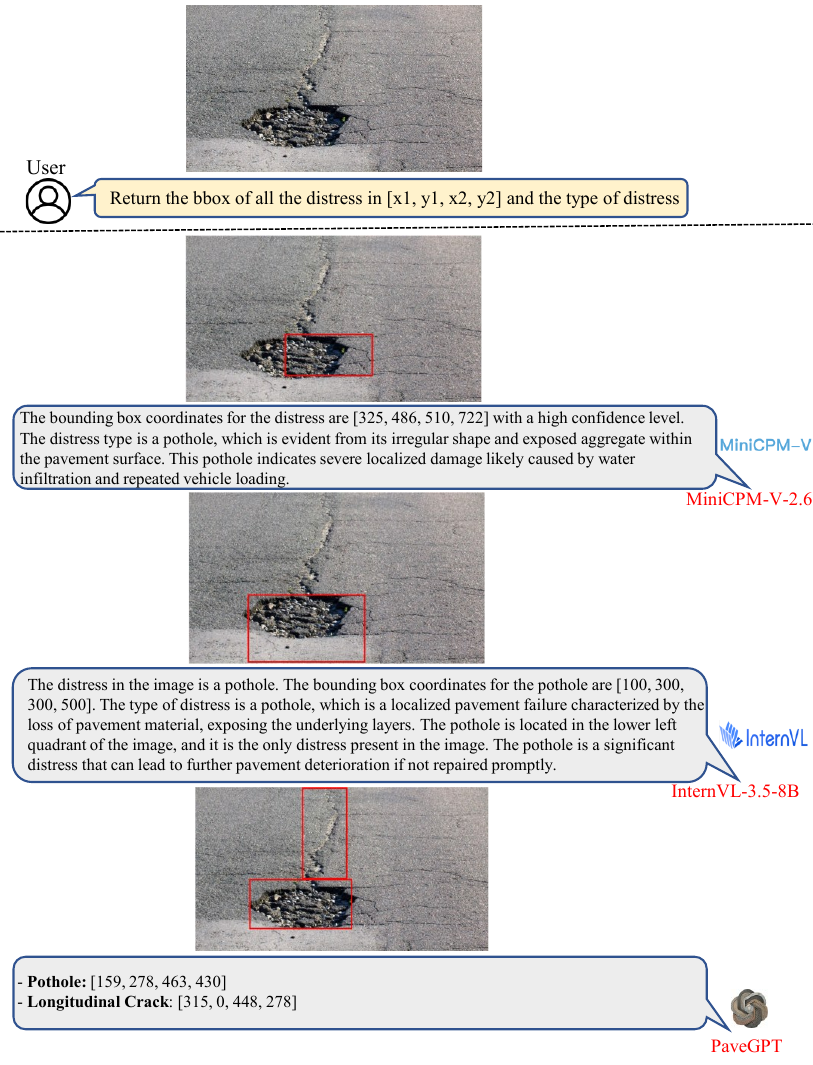}
\caption{Model predictions for spatial grounding task. Red boxes indicate localized distresses.}
\label{fig:spatial_grounding}
\end{figure}

Finally, we evaluate spatial grounding, where models are required to localize visible distresses using bounding boxes. Figure~\ref{fig:spatial_grounding} shows that PaveGPT correctly identifies and localizes both a pothole and a longitudinal crack, producing bounding boxes in the expected [x1, y1, x2, y2] format with clear class labels. InternVL and MiniCPM also detect both objects, but PaveGPT’s boxes exhibit more precise alignment and clearer labeling. Importantly, PaveGPT explicitly associates each bounding box with a distress type, which supports downstream use in mapping, inspection reporting, and asset management systems. These examples demonstrate that instruction tuning improves both visual localization accuracy and structured output formatting.

These qualitative results show that supervised fine-tuning on PaveInstruct leads to more accurate, structured, and interpretable outputs across diverse pavement assessment tasks. The models demonstrate improved alignment with domain terminology, reasoning protocols, and output formats, which are essential for practical deployment in real-world pavement evaluation workflows.

\subsection{Model Efficiency and Complexity}
To assess the computational requirements of each model, we conduct inference benchmarks on an NVIDIA A6000 GPU with 48~GB of memory. We measure four key metrics that characterize model efficiency during deployment, including Time to First Token, token generation throughput, peak memory usage, and overall efficiency in tokens per second per gigabyte (TPS/GB). These metrics reflect real-world usability, particularly in scenarios where speed, memory usage, and computational cost are important. 

\textbf{Time to First Token (TTFT)} measures the latency between submitting an input and receiving the first generated token. It captures the initial processing overhead, including image encoding, instruction parsing, and model warm-up. Lower TTFT values indicate faster responsiveness, which is critical for interactive applications or real-time deployments.

\textbf{Throughput} refers to the average number of tokens generated per second after the first token is produced. This reflects how efficiently the model handles long or multi-turn outputs. Higher throughput implies faster text generation, making the model more suitable for inference at scale.

\textbf{Memory Usage} reports the peak GPU memory consumption during inference. This includes memory allocated for the model weights, attention buffers, and intermediate activations. Lower memory usage enables deployment on edge devices or smaller accelerators and allows concurrent processing of multiple inputs.

\textbf{Efficiency (TPS/GB)} is defined as the number of tokens generated per second per gigabyte of memory used. It combines speed and memory usage into a single metric, allowing direct comparison of how well each model balances performance and resource consumption. Higher values indicate more efficient use of hardware resources.

\textbf{Total Time} is the overall wall-clock time taken to generate a complete response for a single example. This includes both the TTFT and the time spent generating all tokens. While influenced by throughput, it provides an intuitive sense of end-to-end latency from the user's perspective.

Table~\ref{tab:inference} presents the inference benchmarks for all evaluated models. Among the compared models, PaliGemma-3B achieves the lowest TTFT of 60 ms and the highest throughput of 64.3 tokens per second while consuming only 6.02 GB of memory. This results in an efficiency of 10.7 TPS/GB, which is the highest among all models. However, this computational advantage comes at the cost of reduced task performance, as shown in our evaluation results. LLaVA-1.5 demonstrates the second-best latency at 93 ms and achieves a throughput of 44.2 tokens per second with an efficiency of 3.0 TPS/GB, making it a strong choice when balancing speed and capability.

\begin{table}[H]
\centering
\caption{Inference efficiency benchmarks measured on an NVIDIA A6000 GPU (48 GB). TTFT: Time to First Token. Efficiency is computed as Throughput / Memory (TPS/GB). Best results are shown in \textbf{bold}, second best are \underline{underlined}.}
\label{tab:inference}
\resizebox{\columnwidth}{!}{%
\begin{tabular}{lccccc}
\toprule
\textbf{Model} & \textbf{TTFT (ms)} $\downarrow$ & \textbf{Throughput (tok/s)} $\uparrow$ & \textbf{Memory (GB)} $\downarrow$ & \textbf{Efficiency (TPS/GB)} $\uparrow$ & \textbf{Total Time (s)} $\downarrow$ \\
\midrule
PaliGemma-3B & \textbf{60} & \textbf{64.3} & \textbf{6.02} & \textbf{10.7} & \textbf{1.554} \\
LLaVA-1.5-7B & \underline{93} & \underline{44.2} & \underline{14.51} & \underline{3.0} & \underline{2.263} \\
LLaVA-1.6-7B & 347 & 39.0 & 15.71 & 2.5 & 2.566 \\
InternVL-3.5-8B & 307 & 28.7 & 17.58 & 1.6 & 3.481 \\
LLaMA-3.2-11B & 253 & 30.8 & 22.31 & 1.4 & 3.247 \\
PaveGPT-7B & 236 & 39.7 & 16.81 & 2.4 & 2.518 \\
\bottomrule
\end{tabular}%
}
\end{table}

PaveGPT achieves a TTFT of 236 ms and a throughput of 39.7 tokens per second while using 16.81 GB of memory. This results in an efficiency of 2.4 TPS/GB, which is comparable to LLaVA-1.6 (2.5 TPS/GB) despite PaveGPT's superior task performance. The slightly higher memory consumption of PaveGPT relative to LLaVA 1.5 can be attributed to the more complex vision encoder and language model architecture of Qwen2.5-VL. LLaMA 3.2 Vision exhibits the highest memory usage at 22.31 GB and the lowest efficiency at 1.4 TPS/GB, which is expected given its larger 11B parameter count. InternVL3-8B shows similar memory characteristics to PaveGPT but achieves lower throughput at 28.7 tokens per second. These results indicate that PaveGPT provides a reasonable trade-off between inference efficiency and task performance, making it suitable for practical pavement inspection applications where both accuracy and computational feasibility are important considerations.

\section{Practical Applications and Limitations}
The instruction-tuned models enable construction agencies to replace multiple specialized inspection systems with a single unified tool that handles diverse assessment tasks through natural language interaction. Field inspectors conducting building envelope assessments, concrete structure monitoring, or pavement surveys can conversationally query conditions and receive engineering standards-compliant responses immediately. The computational efficiency demonstrated makes deployment on edge devices practical without requiring cloud connectivity, enabling on-site inspection at construction projects and infrastructure assets.

Operational deployment requires fine-tuning on local inspection images to capture region-specific defect patterns and construction practices, then integrating with existing management software through standard APIs. Training programs can use the models as interactive tools that guide new inspectors through defect identification and condition rating procedures for various construction materials and building systems. However, agencies must establish data governance policies for managing inspection images that may capture identifying information.

Several technical limitations affect practical deployment. The dataset reflects conditions from specific geographic regions and construction practices, making local fine-tuning essential for areas with different defect patterns or environmental conditions. Instruction generation relies on automated synthesis validated through expert review, meaning responses approximate rather than replicate experienced inspector communication. Evaluation measures technical performance but not operational factors like inspector trust or workflow integration complexity. Condition estimation errors range from 13 to 31 points depending on architecture, suggesting use as screening tools rather than for final ratings driving capital investment decisions or regulatory compliance.

The focus on static images limits temporal degradation tracking because the dataset lacks annotations linking repeated observations over time. This prevents forecasting capabilities needed for proactive maintenance planning and construction quality monitoring across project phases. These limitations indicate that agencies should validate performance in their construction contexts, treat outputs as decision support augmenting inspector judgment, and focus initial deployments on preliminary screening where current capabilities align with operational needs.

\section{Conclusion}

This work addresses the challenge of applying vision-language models to specialized technical domains through instruction-driven pavement condition assessment. While general-purpose VLMs perform well on everyday tasks, they struggle with domain-specific requirements such as technical terminology, structured reasoning, and adherence to engineering standards. We introduce \emph{PaveInstruct}, a comprehensive dataset of 278,889 instruction-response pairs spanning 32 task types, from spatial grounding to PCI estimation and maintenance recommendations. The dataset integrates annotations from nine heterogeneous sources through a systematic unification pipeline that preserves semantic richness while ensuring engineering validity. We also
present \emph{PaveGPT}, a domain-specialized foundation model trained on this
dataset that achieves strong performance across perception, understanding,
and reasoning tasks.

Our experiments show that domain-specific instruction tuning fundamentally transforms model capabilities. Models that failed in zero-shot settings achieved improvements of over 20\% in spatial grounding, reasoning, and generation tasks after fine-tuning. More importantly, fine-tuned models produced technically accurate outputs aligned with ASTM standards, generated well-structured responses suitable for pavement management systems,
and demonstrated professional reasoning chains. These improvements were consistent across different architectures, indicating that targeted supervision
rather than model scale drives performance gains.

The practical value extends beyond pavement assessment. Our approach enables conversational interfaces where professionals can query conditions, request maintenance priorities, and obtain justified assessments through natural language. This unified framework replaces multiple specialized models
with a single instruction-following system that handles diverse workflows. However, our dataset reflects specific geographic and climatic conditions that may not generalize to all regions, and PCI estimation errors may exceed tolerances for certain regulatory applications. Future work should expand PaveInstruct to additional regions and pavement types, develop real-time processing pipelines integrated with sensor data, and test deployment in operational management systems. Extending this approach to other infrastructure domains such as bridge inspection would validate its broader applicability.

\section{Acknowledgements}
The authors gratefully acknowledge the use of the iTiger HPC cluster, operated by the University of Memphis Research Computing group, which provided access to NVIDIA H100 GPUs.



\bibliographystyle{elsarticle-num}
\bibliography{export}
\end{document}